\documentclass{article}

\newif\ifconference

\ifconference
\usepackage{iclr2026_conference, times}
\PassOptionsToPackage{numbers, sort&compress}{natbib}
\else
\usepackage[numbers,sort&compress]{natbib}
\usepackage{fullpage}
\usepackage{authblk}
\usepackage{lmodern}
\usepackage{setspace}
\fi

\usepackage{amsmath,amsfonts,bm,amsthm}

\def\eqref#1{equation~\ref{#1}}

\def\1{\bm{1}}

\def\eps{{\epsilon}}

\DeclareMathAlphabet{\mathsfit}{\encodingdefault}{\sfdefault}{m}{sl}
\SetMathAlphabet{\mathsfit}{bold}{\encodingdefault}{\sfdefault}{bx}{n}

\newcommand{\R}{\mathbb{R}}

\DeclareMathOperator*{\argmin}{arg\,min}

\usepackage{enumitem}
\usepackage[utf8]{inputenc} 
\usepackage[T1]{fontenc}    
\usepackage{hyperref}       
\usepackage{url}            
\usepackage{booktabs}       
\usepackage{amsfonts}       
\usepackage{nicefrac}       
\usepackage{microtype}      
\usepackage{xcolor}         

\theoremstyle{plain}
\newtheorem{theorem}{Theorem}[section]
\newtheorem{proposition}[theorem]{Proposition}

\theoremstyle{definition}
\newtheorem{definition}[theorem]{Definition}
\newtheorem{assumption}[theorem]{Assumption}
\theoremstyle{remark}

\newcommand{\dd}{\mathrm{d}}

\usepackage{amsmath,amssymb,amsthm,latexsym,color,mathtools} 
\usepackage[dvipsnames]{xcolor}
\usepackage{braket} 

\definecolor{MIT}{cmyk}{.24, 1.00, .78, .17}

\allowdisplaybreaks[2] 
\let\temp\phi
\let\phi\varphi
\let\varphi\temp

\let\temp\epsilon
\let\epsilon\varepsilon
\let\varepsilon\temp

\renewcommand{\hat}{\widehat}

\newcommand{\Z}{{\mathbb Z}}

\newcommand{\Acl}{\mathcal{A}}

\newcommand{\Dcl}{\mathcal{D}}
\newcommand{\Ecl}{\mathcal{E}}
\newcommand{\Fcl}{\mathcal{F}}

\newcommand{\Ncl}{\mathcal{N}}

\newcommand{\Zcl}{\mathcal{Z}}

\renewcommand{\epsilon}{\varepsilon}

\newtheoremstyle{plain}
  {3pt}   
  {0pt}   
  {\itshape}  
  {0pt}       
  {\bfseries} 
  {.}         
  {3pt} 
  {}          

\newtheoremstyle{exercise}
 {3pt}
 {3pt}
 {}
 {}
 {\bfseries}
 {.}
 {3pt}
 {\thmname{#1} \thmnumber{#2}, \thmname{#3}}
 
\newtheoremstyle{claim}
 {3pt}
 {3pt}
 {}
 {}
 {\bfseries}
 {.}
 {3pt}
 {\thmname{#1} \thmnumber{#2}}
 
\theoremstyle{exercise}

\theoremstyle{plain}
\newtheorem{-thm}{Theorem}[section]
\newtheorem{-prop}[-thm]{Proposition}
\newtheorem{-lem}[-thm]{Lemma}
\newtheorem{-cor}[-thm]{Corollary}

\theoremstyle{definition}

\theoremstyle{claim}

\makeatletter
\renewenvironment{proof}[1][\proofname]{
  \pushQED{\qed}%
  \normalfont \partopsep=\z@skip \topsep=\z@skip
  \trivlist
  \item[\hskip\labelsep
        \itshape
    #1\@addpunct{.}]\ignorespaces
}{%
  \popQED\endtrivlist\@endpefalse \medskip
}
\makeatother

\usepackage{hyperref}
\usepackage{url}
\usepackage{wrapfig}

\newcommand{\authrone}{Mara Daniels}
\newcommand{\affilone}{Department of Mathematics}
\newcommand{\addrsone}{Massachusetts Institute of Technology}
\newcommand{\emailone}{maradan@mit.edu}
\newcommand{\webone}{\url{https://maradan.me/}}

\newcommand{\authrtwo}{Liam Hodgkinson}
\newcommand{\affiltwo}{School of Mathematics and Statistics}
\newcommand{\addrstwo}{University of Melbourne}
\newcommand{\emailtwo}{lhodgkinson@unimelb.edu.au}
\newcommand{\webtwo}{\url{https://www.liamhodgkinson.com}}

\newcommand{\authrthree}{Michael W. Mahoney}
\newcommand{\affilthree}{ICSI \\ LBNL \\ and Department of Statistics}
\newcommand{\addrsthree}{University of California at Berkeley}
\newcommand{\emailthree}{mmahoney@stat.berkeley.edu}
\newcommand{\webthree}
{\url{https://www.stat.berkeley.edu/~mmahoney}}

\title{Uncertainty-Aware Diagnostics for \\ Physics-Informed Machine Learning}

\ifconference
\author{%
  \authrone\thanks{\webone} \\
  \affilone\\
  \addrsone\\
  \texttt{\emailone} \\
  \And
  \authrtwo\thanks{\webtwo} \\
  \affiltwo \\
  \addrstwo \\
  \texttt{\emailtwo}\\
  \AND
  \authrthree\thanks{\webthree} \\
  \affilthree \\
  \addrsthree \\
  \texttt{\emailthree} \\
}
\else
    \author[1]{\authrone}
    \author[2]{\authrtwo}
    \author[3]{\authrthree}
    \affil[1]{\small{\textit{\affilone, \addrsone} \protect\\ \href{mailto:\emailone}{\protect\nolinkurl{\emailone}}} \vspace{2mm}}
    \affil[2]{\small{\textit{\affiltwo, \addrstwo} \protect\\ \href{mailto:\emailtwo}{\protect\nolinkurl{\emailtwo}}} \vspace{2mm}}
    \affil[3]{\small{\textit{\affilthree, \addrsthree} \protect\\ \href{mailto:\emailthree}{\protect\nolinkurl{\emailthree}}} \vspace{2mm}}
    \date{}
\fi

\begin{document}

\maketitle

\begin{abstract}

Physics-informed machine learning (PIML) integrates prior physical information, often in the form of differential equation constraints, into the process of fitting machine learning models to physical data. 
Popular PIML approaches, including neural operators, physics-informed neural networks, neural ordinary differential equations, and neural discrete equilibria, are typically fit to objectives that simultaneously include both data and physical constraints. 
However, the multi-objective nature of this approach creates ambiguity in the measurement of model quality.
This is related to a poor understanding of epistemic uncertainty, and it can lead to surprising failure modes, even when existing statistical metrics suggest strong fits. 
Working within a Gaussian process regression framework, we introduce the Physics-Informed Log Evidence (PILE) score. 
Bypassing the ambiguities of test losses, the PILE score is a single, uncertainty-aware metric that provides a selection principle for hyperparameters of a PIML model. 
We show that PILE minimization yields excellent choices for a wide variety of model parameters, including kernel bandwidth, least squares regularization weights, and even kernel function selection. 
We also show that, even prior to data acquisition, a special ``data-free'' case of the PILE score identifies \emph{a priori} kernel choices that are ``well-adapted'' to a given PDE. 
Beyond the kernel setting, we anticipate that the PILE score can be extended to PIML at large, and we outline approaches to do so.

\end{abstract}

\section{Introduction}

\ifconference
\vspace{-.2cm}
\else
\fi

A great challenge in machine learning (ML) in general and Scientific ML (SciML) in particular involves the development of models that can combine in principled ways data-driven information (as is common in ML) and domain-driven information (as is common in physical and other sciences). 
Strategies that attempt to achieve this are grouped under the umbrella of \emph{physics-informed machine learning} (PIML).
These approaches consist of general purpose tools for scientific computation that also enjoy the scalability and flexibility of high-dimensional ML.  
PIML methods include \emph{Physics-Informed Neural Networks} (PINNs) \citep{raissi2019physics,krishnapriyan2021characterizing} (which led to a large body of empirical work \citep{krishnapriyan2021characterizing,karniadakis2021physics,sirignano2018dgm,sahli2020physics,jin2021nsfnets,geneva2020modeling,xu2020physics} and theoretical work \citep{mina2023apriori,lu2021apriori,pmlr-v247-doumeche24a}), \emph{Neural Ordinary Differential Equations} (Neural ODEs) \citep{chen2018neural, krishnapriyan2023learning}, \emph{Neural Operators} \citep{kovachki2023neural,FalsePromizeZeroShot_TR}, and \emph{Neural Discrete Equilibrium} (NeurDE) \citep{neurde_TR}.
Unfortunately, PINNs and related methods are notoriously difficult to train \citep{krishnapriyan2021characterizing}; they lack robust \emph{a posteriori} error estimates that are typically available for classical numerical partial differential equation (PDE) solvers; and they lack a strong grounding in statistical theory. 
These issues are exacerbated by the multi-objective nature of PINNs, as well as other PIML methods that incorporate domain knowledge as a soft regularization. 

It is (easy and thus) popular to view physical constraints in terms of regularizers, in a manner analogous to ridge regression \citep{karniadakis2021physics}.
However, practical implementations involve a delicate trade-off between errors to noisy observations and adherence to the imposed (or assumed) physical equations.
Indeed, one interpretation of the ``failure modes'' results of \citet{krishnapriyan2021characterizing} is that while reducing physical error is of interest, this strategy becomes nuanced (and error-prone) when the model is unable to satisfy the physical constraints perfectly, or when the constraints are misspecified. 
In these cases, without sufficient validation data (which is common in scientific settings), it becomes challenging to determine whether a model is a suitable fit.
These challenges are not limited to neural network models---they are common to many other methods that aim to combine data-driven ML models with domain-driven physical models.
\textbf{It is thus critical to understand how to quantify the quality of PIML models}, in a manner analogous to how one quantifies model quality in statistical learning theory. 

In this paper, we provide a first step toward solving this PIML model selection problem, addressing the problem under the Physics-Informed Kernel Learning (PIKL) framework \citep{pförtner2024physicsinformedgaussianprocessregression}.
PIKL considers a Gaussian process (GP) model for solving linear PDEs under known conditions, and it offers a powerful, uncertainty-aware approach to PIML.
One advantage of the GP framework is that it offers a structured, probabilistic approach that allows for rigorous uncertainty quantification (UQ) (which is often missing in other physics-informed models\footnote{Some exceptions exist, including Neural Processes \citep{garnelo2018neural,kim2019attentive}, but they will prove to be disadvantageous in our framework due to intractable marginal likelihoods.}). 
Another advantage of GPs lies in their ability to seamlessly incorporate multiple forms of data acquisition, including noisy pointwise observations of the solution and derivative data derived from the governing PDE. 
This flexibility enables a principled integration of prior knowledge about the system, while maintaining a Bayesian framework for uncertainty estimation. 

Our main contributions are as follows:
\begin{enumerate}[leftmargin=*,label=(\Roman*)]
\item 
We introduce a \textbf{model selection criterion} called the \textit{Physics-Informed Log Evidence (PILE)} (Section~\ref{sec:construction}). 
As with other statistical model selection criteria, the PILE criterion is based on the free energy, and it provides a theoretically-grounded way to assess the suitability of different kernel choices for PIML tasks; and it can be used, e.g., to optimize hyperparameters of the GP model, including the kernel function, its bandwidth, and regularization parameters.
\item 
We provide an \textbf{empirical evaluation} demonstrating that the PILE criterion is a reliable indicator of model performance (Section \ref{sec:empirical}). 
Models optimized using PILE exhibit strong predictive accuracy and adherence to physical constraints. By studying the challenging wave-equation setting introduced in \citet{krishnapriyan2021characterizing}, we show that the PILE score can not only diagnose model misspecification, but it can be used to identify the ``best'' kernel function for the problem at hand, leading to vastly improved performance.
\end{enumerate}

Overall, \textbf{we claim that free energy metrics provide the solution to the multi-objective bottleneck in PIML}, establishing a single number that can be optimised to ensure \emph{both} a strong fit to existing data \emph{and} adherence to a governing differential equation. 
Diagnostics using the free energy can be conducted both \emph{a priori} (at the stage of architecture selection and before fitting the model to data), 
using a special ``data-free'' case of our PILE score
(which is related to a Fredholm determinant), 
and also \emph{a posteriori} 
(after the model has been fitted to data), 
using our PILE score. 
Our case studies demonstrate how the use of these tools can bypass well-known pitfalls in PIML, highlighting scenarios where a model choice will lead to an undesirable fit. %

\ifconference
\vspace{-.2cm}
\else
\fi

\section{Background and Related Work}
\ifconference
\vspace{-.3cm}
\else
\fi

\subsection{Linear Partial Differential Equations}
\ifconference
\vspace{-.2cm}
\else
\fi

An $s$-th order linear differential operator $\Dcl : C^{s}(\Omega) \to C^0(\Omega)$ for integer $s \geq 1$ has the form 
\begin{align*}
    \Dcl f(x) = \sum_{\|\alpha\|_1 \leq s} c_\alpha(x) \frac{\partial^{\alpha_1}}{\partial x_1^{\alpha_1}} \cdots \frac{\partial^{\alpha_d}}{\partial x_d^{\alpha_d}} f(x)  ,
\end{align*}
where $\Omega$ is an open, bounded subset of $\R^d$ with $C^1$ boundary, where $\alpha \in \Z_{\geq 0}^d$ is a multi-index, and where $\{c_\alpha : \|\alpha\|_1 \leq s\}$ are $C^0(\Omega)$ coefficient functions. Given $g \in C^0(\Omega)$ and $h \in C^0(\partial \Omega)$, we say that $f$ solves the \textit{Dirichlet boundary value problem} (BVP) if 
\begin{align}\label{eqn:dbvp}
    \Dcl f(x) = g(x)\; \text{ for } x \in \Omega,\qquad  
        f(x) = h(x)\; \text{ for } x \in \partial \Omega .
\end{align}
Under our assumptions, $f$ solves the Dirichlet BVP if and only if it minimizes the energy functional 
\begin{align} 
\label{eqn:energy-func}
    \Ecl(f) \coloneqq \|\Dcl f - g\|_{L^2(\Omega)}^2 + \|f - h\|_{L^2(\partial \Omega)}^2  .   
\end{align}
This variational problem is a key ingredient in the formulation of \eqref{eqn:dbvp} as a ML problem. 
Note, however, that we can extend our formulation far beyond the Dirichlet setting to encompass a wide range of mixed boundary conditions. 
To simplify matters, observe that the formulation is no less general if $g = 0$, as we can instead take the differential operator $\mathcal{D} - g \mathsf{Id}$. 
Let $\mathcal{D}$ be as before, but now let $\mathcal{B}_i:C^{s}(\Omega)\to C^{0}(\Gamma_i)$, $i=1,\ldots, p$ be a family of operators, where each $\Gamma_{i}\subset\partial\Omega$. We can now consider the general mixed boundary condition
\[
\mathcal{D}f(x)=0\;\text{ for }x\in\Omega,\qquad(\mathcal{B}_if)(x)=0\;\text{ for } x \in \Gamma_i, \ i=1,\ldots, p  .
\]
This formulation can encode the following boundary conditions:
\begin{itemize}[leftmargin=*]
\item \textbf{Dirichlet:} $p=1$, $\Gamma_{1}=\partial\Omega$, and $\mathcal{B}_1f=(f-h)\vert_{\partial\Omega}$ is the restriction of f to~$\partial\Omega$.
\item \textbf{Neumann:} $p=1$, $\Gamma_{1}=\partial\Omega$, and $\mathcal{B}_1f=(\nu\cdot\nabla f-h)\vert_{\partial\Omega}$, where $\nu$ is the unit normal to $\partial\Omega$.
\item \textbf{Robin:} $p=1$, $\Gamma_{1}=\partial\Omega$, and $\mathcal{B}_1f=(a f+b\nu\cdot\nabla f-h)\vert_{\partial\Omega}$ for $a, b \in \mathbb{R}$.
\item \textbf{Cauchy:} $p=2$, $\Gamma_{1}=\Gamma_{2}=\partial \Omega$, $\mathcal{B}_1 f= (f-h_{1})\vert_{\Gamma_{1}}$, and $\mathcal{B}_2 f = (\nu\cdot\nabla f-h_{2})\vert_{\Gamma_{2}}$. 

\end{itemize}
In these cases, any solution $f$ is a minimizer of the energy functional
\begin{equation}
\label{eq:EnergyGen}
\mathcal{E}(f) \coloneqq \|\Dcl f\|_{L^2(\Omega)}^2 + \sum_{i=1}^p \|\mathcal{B}_i f\|_{L^2(\Gamma_i)}^2.
\end{equation}
Letting $\Acl f(x)= (\Dcl f(x)\boldsymbol{1}_{\Omega}, \mathcal{B}_1 f(x)\boldsymbol{1}_{\Gamma_1}, \ldots, \mathcal{B}_p f(x)\boldsymbol{1}_{\Gamma_p}) \in \R^{p+1}$, we can define a measure $\mu$ that is Lebesgue on $\Omega$ and Hausdorff on each $\Gamma_i$ so that $\mathcal{E}(f) = \|\Acl f\|_{L^2(\bar{\Omega},\mu,\mathbb{R}^{p+1})}^2$. This notation will become convenient later. While integral constraints can also be incorporated naturally within this setup \citep{hansen2023learning}, here we restrict attention to differential operator constraints to avoid complicating our analysis.

\ifconference
\vspace{-.2cm}
\else
\fi

\subsection{Gaussian Process Regression} 
\label{sec:GPRegression}
\ifconference
\vspace{-.1cm}
\else
\fi

A \emph{Gaussian process (GP)} $f$ on $\Omega \subseteq \mathbb{R}^d$, denoted $f \sim \mathcal{GP}(m, k)$, is a stochastic process where for some \emph{mean function} $m : \Omega \to \R$ and a \textit{kernel function} $k : \Omega \times \Omega \to [0,\infty)$, any projection onto finitely many points $X = \{x_i\}_{i=1}^n \subseteq \Omega$ is multivariate Gaussian:
$$
( f(x_1), \dots, f(x_n) ) \sim \Ncl((m(x_i))_{i=1}^n, (k(x_i, x_j))_{i,j=1}^n).
$$
As covariance matrices are necessarily symmetric positive semi-definite, we require that the \emph{Gram matrix} $(k(x_i, x_j))_{i,j=1}^n$ is a positive semi-definite matrix, for any $\{x_i\}_{i=1}^n$. 
(Any function $k$ with this property is said to be positive semi-definite.) We further require $k$ to be continuous on $\Omega \times \Omega$ and to have $\int_{\Omega} k(x,x) \dd x < \infty$. 

GP regression is a framework which provides a Bayesian perspective on kernel regression, along with a probabilistic interpretation of the commonly-used kernel ridge regularization. 
Given independent and identically distributed inputs $x_i \in \R^d$ and outputs $y_i \in \R$, for $i=1\ldots n$, a Gaussian likelihood
\[
y_i \mid (f, x_i) \sim \Ncl(f(x_i), \tfrac12 \gamma),\quad p(y_i \mid f, x_i) \propto \exp(-\tfrac{1}{\gamma}(y_i - f(x_i))^2),\quad i=1,\dots,n,
\]
is imposed, where $\gamma > 0$ is a hyperparameter representing the assumed noise level of the observations. 
In the noise-free setting, one can take $\gamma \to 0^+$. 
For notational convenience, let $X = (x_{ij})_{i,j=1}^{n,d} \in \mathbb{R}^{n\times d}$ and $Y = (y_i)_{i=1}^n \in \mathbb{R}^n$. 
To apply Bayes' theorem, the practitioner chooses a GP prior, $f \sim \mathcal{GP}(m, \lambda^{-1} k)$, for $\lambda >0$ a regularization hyperparameter. 
In the absence of prior information, it is common to choose $m \equiv 0$. 
To perform inference, the prediction and uncertainty for the output of a new input $x'$ is measured using the posterior predictive distribution \citep[Equation 2.19]{Rasmussen2006Gaussian}
\[
f(x') \mid (x_1,y_1),\dots,(x_n,y_n) \sim \Ncl(\bar{f}(x'), \lambda^{-1} \sigma(x')),
\]
where $\bar{f}(x) = k_{x'}^\top (K_X + \lambda \gamma I)^{-1} Y$ and $\sigma(x') = k(x',x') - k_{x'}^\top (K_X + \lambda \gamma I)^{-1} k_{x'}$ for $k_{x'}~=~(k(x_i,x'))_{i=1}^n$ and $K_X~=~(k(x_i,x_j))_{i,j=1}^n$. 
In addition to the point estimate, $\bar{f}(x)$, the posterior variance, $k_{x'}$, can be taken as a calibrated measure of predictive uncertainty. 

Kernel ridge regression (KRR) is a prediction method (with limited UQ) that estimates the output $f(x)$ for a given input $x$ by minimizing a regularized loss over a reproducing kernel Hilbert space (RKHS). 
Recall that every positive-definite kernel $k$ induces a RKHS, $H$, generated by the span of $\{k(x,\cdot)\}_{x \in \Omega}$ with norm $\|\cdot\|_H$ (see Appendix \ref{sec:RKHS} for details). 
KRR considers estimators of the form:
\begin{equation}
\label{eq:MAP}
\hat{f} = \argmin_{f \in H} \frac1\gamma \sum_{i=1}^n (f(x_i) - y_i)^2 + \lambda \|f\|_{H}^2.
\end{equation}
From \citet[Theorem 3.4]{kanagawa2018gaussian}, it turns out that $\hat{f} = \bar{f}$, the mean predictor from the GP formulation, and so GP regression extends KRR to include an estimate of the uncertainty. Simultaneously, any optimization problem of the form (\ref{eq:MAP}) has a natural interpretation in terms of an underlying GP, providing uncertainty estimates for solutions to (\ref{eq:MAP}). We will make significant use of this relationship to construct our uncertainty-aware diagnostics.

\ifconference
\vspace{-.2cm}
\else
\fi

\subsection{Uncertainty and Diagnostics in GP}
\ifconference
\vspace{-.15cm}
\else
\fi

One of the advantages with treating a prediction task through the lens of statistical models lies in the ability to analyze and estimate uncertainty. 
Bayesian methods, including GPs, naturally account for prediction uncertainty in the \emph{posterior predictive distribution}, offering credible intervals for any quantile of uncertainty \citep{gelman1995bayesian}. 
For GPs, the uncertainty about the prediction is contained in the posterior covariance kernel $x \mapsto \Sigma(x,x)$.
Estimates of uncertainty are only as effective as the underlying model. Fortunately, the treatment of uncertainty often unlocks a wide array of diagnostic techniques, providing valuable feedback to the practitioner. 

In Bayesian statistics, the fundamental indicator of the quality of a particular prior is the \textit{marginal likelihood}, also known as the \textit{evidence}, given by \vspace{-.3cm}
\begin{align}\label{eqn:evidence}
    \Zcl_n = \mathbb{E}_{f \sim \mathcal{GP}(0, \lambda^{-1}k)}[p(Y \mid f, X)].
\end{align}
This quantity can be thought of as the likelihood assigned by the prior to the observed data. 
It is typically convenient to work instead with the quantity
\begin{align}\label{eqn:nmll}
    \Fcl_n \coloneqq - \log \Zcl_n  = \frac{1}{2} Y^\top(K_X + \gamma I)^{-1} Y  + \frac{1}{2} \log \det (K_x + \gamma I) - \frac{n}{2} \log \left(\frac{\lambda}{2 \pi} \right)  ,
\end{align}
which is called the (negative) \textit{log-marginal likelihood}, alternatively the \textit{Bayes free energy}. 
In practice, it is common to perform model selection by maximizing the marginal likelihood (equivalently, minimizing the free energy $\Fcl_n$) with respect to hyperparameters of the prior. 
This is called an \textit{empirical Bayes procedure} \citep{krivoruchko2019evaluation}, and it is the main inspiration of the PILE score which we introduce in Section \ref{sec:construction}. 

Aside from model selection, the free energy is also effective for model tuning. 
This process is referred to as \emph{empirical Bayes} \citep{efron2024empirical}; and, provided that not too many parameters are tuned this way, it is often effective \citep{lotfi2022bayesian}. 
For GPs, bandwidth tuning in the kernel, and the selection of the noise level $\gamma$, are both often conducted by minimizing Bayes free energy \citep[5.4.1]{Rasmussen2006Gaussian}; see also \citet{gribov2020empirical}. 
However, information criteria---especially Bayesian ones---are not equivalent to test or cross-validation error.
This leads to the question: when do they behave similarly? 
Fortunately, for GPs, it is known that the Bayes free energy behaves similarly to test error---at least when the number of training and test points are large; see, e.g., \citet{pmlr-v202-hodgkinson23a,luxburg2004distance,jinlearning}.

\ifconference
\vspace{-.2cm}
\else
\fi

\section{Physics-Informed Kernel Learning}
\label{sec:construction}
\ifconference
\vspace{-.2cm}
\else
\fi

The GP formulation of PIKL is based on a finite sample approximation of \eqref{eqn:energy-func}, using (possibly noisy) observations of the graph $(f, \Acl f)$ at points $x \in \Omega$. 
Following \citet{doumeche2024physics}, we consider minimizers of the physics-informed empirical risk over $f$ in a RKHS $H$:
\begin{equation}
\label{eq:MainLoss}
L_n(f) \coloneqq \frac{1}{\gamma}\cdot  \underset{\textrm{data loss}}{\underbrace{\frac{1}{n}\sum_{i=1}^n (f(x_i) - y_i)^2}} + \frac{1}{\rho} \underset{\textrm{physics loss}}{\underbrace{\vphantom{\sum_i}\|\mathcal{A} f\|^2_{L^2(\bar{\Omega},\mu,\mathbb{R}^{p+1})}}} + \frac{1}{\eta} \underset{\textrm{regularization}}{\underbrace{\vphantom{\sum_i}\|f\|_H^2}}.
\end{equation}
The first term is the \emph{data loss}, prescribing adherence of $f$ to collected observations in the form of input-output pairs $(x_i,y_i)$, with $x_i \in \bar{\Omega}$ and $y_i \in \mathbb{R}$. These can be used in addition to, or in place of, boundary conditions. 
The second term is the \emph{physics loss}, which enforces that $f$ obey the prescribed equation $\mathcal{A} f = 0$. 
The third term biases the estimator towards a \emph{more regular solution}, avoiding spikes and other singular behavior. 
For example, the RKHS $H_\nu$ associated with the Matern kernel of smoothness parameter $\nu > 0$ is equivalent to the Sobolev space $W^{\nu+d/2,2}(\Omega)$ \cite[Corollary 10.13]{Wendland_2004}, and so we have $\|f\|_{H_{\nu}(\bar{\Omega})}^2 \asymp \|f\|_{W^{\nu+d/2,2}(\bar{\Omega})}^2$, the sum of the $L^2$ norm of the first $s$ derivatives of $f$.  
The temperatures $\gamma,\rho,\eta > 0$ are arbitrary and control the relative importance of these three terms. Since scaling $L^{\eta,\rho,\gamma}$ does not change its minimizer, it is typical to fix one parameter and vary the other two. For quantifying the uncertainty in model predictions pointwise in space, however, selecting the correct scale of $L^{\eta, \rho, \gamma}$ is required to have accurate and calibrated estimates of the pointwise posterior variance, and thus all three parameters are needed.

One approach to solving (\ref{eq:MainLoss}) using KRR, taken in \citet{doumeche2024physics,pmlr-v247-doumeche24a}, is to identify a new RKHS $H'$ with norm $\|f\|_{H'}^2 = \|f\|_H^2 + \frac{\eta}{2\rho} \|\Acl f\|_{L^2(\bar{\Omega},\mu,\mathbb{R}^{p+1})}^2$. 
Our approach is different, and it relies on the observation that for a GP $f \sim \mathcal{GP}(m, k)$ over $\Omega$ supported on a Banach space $B \ni f : \Omega \to \R$, the pushforward of the process by a bounded linear operator $\Acl : B \to B'$ is itself a GP, supported on $B'$, with parameters 
$$  \Acl f \sim \mathcal{GP}(\Acl m, (\Acl \otimes \Acl)k).  $$ 
This observation has been applied numerous times in the literature to enforce PDE or other linear constraints on a GP via conditioning on the value of $\Acl f$ \citep[Lemma 2.1]{harkonen2023gaussian}, \citep[Corollary 2]{pförtner2024physicsinformedgaussianprocessregression}, \citep{macêdo2010learning}, \citep{solin2018modeling}. 
While optimal in theory, the corresponding kernel of this space requires deep knowledge of the operator $\mathcal{A}$ and its eigenspectrum. 
Furthermore, samples drawn from a GP can be drastically less regular than the functions contained in the RKHS associated with its covariance kernel.\footnote{This can be seen by the fact that if the Cameron-Martin space associated with $\mathcal{GP}(m, k)$ is infinite dimensional, then it has zero Gaussian measure \citep[Thoerem 3.5.1]{bogachev1998gaussian}. In other words, with probability one, $f \sim \mathcal{GP}(m, k)$ has $\|f\|_H = \infty$ whenever $H$ is not finite dimensional.} 

To avoid this technical issue and to weaken our required assumptions on $k$, one can approximate \eqref{eqn:energy-func} by a physics-informed version of KRR by estimating the $L^2$ norm using a quadrature rule $\{(w_i, z_i)\,:\,i=1,\dots,m\}$ for $\mu$ on $\bar{\Omega}$. One option is a Monte Carlo rule that selects $z_i$ uniformly at random over $\Omega$ and each $\Gamma_i$ with equal weighting $w_i = m^{-1}$. For improved precision, we opt for Gaussian quadrature rules. 
In this case, our estimated loss function becomes
\begin{equation}
\label{eq:ApproxLoss}
L_{m,n}(f) \coloneqq \frac{1}{\gamma n}\sum_{i=1}^n (f(x_i) - y_i)^2 + \frac{1}{\rho} \sum_{i=1}^m w_i (\mathcal{A} f(z_i))^2 + \frac{1}{\eta} \|f\|_H^2.
\end{equation}
The advantage of this approach is that, provided we can formulate $\{(f,\mathcal{A} f)\,:\,f \in H\}$ as a RKHS, (\ref{eq:ApproxLoss}) can be solved using the representer theorem. Let $H$ be a fixed RKHS with reproducing kernel $k$. For a multi-index $\alpha \in \Z_+^d$, we denote by $\partial_1^\alpha k(x, x')$ and $\partial_2^\alpha k(x, x')$ the iterated partial derivative of the first argument and second arguments, respectively: $\partial^\alpha_1 \partial^\beta_2k(x,x') = \partial_{x_1}^{\alpha_1} \,  \cdots \, \partial^{\alpha_d}_{x_d} \, \partial_{x'_1}^{\beta_1}\  \cdots \ \partial_{x_d'}^{\beta_d} k(x, x')$,
where $\partial_{x_i}^{\alpha_i}$ denotes the $\alpha_i$-th partial derivative in $x_i$. 

To proceed, we require 
assumptions on $k$ and $\Acl$. 
\begin{assumption}[Kernel Differentiability]\label{assn:diff} Assume that $k$ has continuous $s$-th partial derivative, that is, for any multi-index $\alpha \in \Z_{\geq 0}^d$, $\|\alpha\|_1 \leq s$, $\partial_x^{\alpha} \partial_{x'}^{\alpha} k(x, x') \in C^0(\Omega \times \Omega)$. 
\end{assumption}
\begin{assumption}[Bounded Coefficients]\label{assn:bdd} 
    Assume that the coefficients $\{c_\alpha : \alpha \in \Z_{\geq 0}^d\}$ for $\mathcal{D}$ and each $\mathcal{B}_i$ are all uniformly bounded as $\max_{\|\alpha\|_1 \leq s} \|c_\alpha\|_{L^\infty(\bar{\Omega})} \leq C < \infty$ for some $C > 0$. 
\end{assumption}
By Proposition \ref{prop:DiffRKHS} in Appendix \ref{sec:RKHS}, under these assumptions, the RKHS associated with $k$ is contained in the image of $\Acl$. Indeed, as in Proposition \ref{prop:NewRKHS}, the graph $\{(f,\mathcal{A}f)\,:\, f\in H\}$ itself an RKHS. Using this fact, in Theorem \ref{prop:representer-theorem}, we obtain a representer theorem which can be used to solve (\ref{eq:ApproxLoss}). Theorem \ref{prop:representer-theorem} is the basis for our variant of the PIKL procedure. Given input-output pairs $\{(x_i,y_i)\}_{i=1}^n$ and a quadrature rule $\{(w_j,z_j)\}_{j=1}^m$, a solution to (\ref{eq:ApproxLoss}), and therefore an approximate solution to (\ref{eq:MainLoss}), can be obtained by finding the coefficients $(\alpha,\beta)$ that minimize (\ref{eqn:representer-theorem}) (in Appendix \ref{sec:RKHS}). 
This procedure is general and particularly effective when $n$ and $m$ are not too large \citep{doumeche2024physics}.

\ifconference
\vspace{-.2cm}
\else
\fi

\section{Physics-Informed Log Evidence (PILE)}
\label{sec:PILE}
\ifconference
\vspace{-.2cm}
\else
\fi

Now that we have demonstrated how PIML problems can be posed in terms of KRR problems, we consider the interpretation of this finding in terms of GP regression, which encodes a natural notion of uncertainty. As discussed in Section \ref{sec:GPRegression}, the estimator $\hat{f}$ of Theorem \ref{prop:representer-theorem} coincides with the posterior predictive mean of a GP model. It can be verified that this model is prescribed by
\begin{equation}
\label{eq:GP}
    (f, g)
    \sim \mathcal{GP}(0, \eta k_{\Acl}),\qquad 
    y_i \mid f(x_i) \sim \Ncl(f(x_i), \tfrac12\gamma), \qquad 
    r_j \mid g(z_j) \sim \Ncl(g(z_j), \tfrac12\rho w_j),
\end{equation}
where each $r_j$ is interpreted as an observation of a boundary condition, which can be taken to be zero to enforce the constraint $\mathcal{A}f = 0$. 
Our formulation targets situations where a practitioner may also have potentially noisy access to the boundary data and forcing function in \eqref{eqn:dbvp}. In this case, one might take $r_j$ to be nonzero. Our goal is to remain as flexible as possible with respect to modes of data acquisition (i.e., possibly corrupted boundary data, interior observations of $f(x)$ or $\Dcl f(x)$ at arbitrary $x \in \bar{\Omega}$) as well as with respect to the model used to represent $f$. 
This flexibility is a key benefit of the PIML approach. 

The uncertainty in the prediction $\hat{f}$ is encoded in the covariance of this GP: for $x',z' \in \bar{\Omega}$,
\[
\mathrm{Cov}(f(x'),\mathcal{A}f(z')) = \eta\left(k_{\Acl}((x',z'),(x',z')) - \varsigma_{x',z'}^\top\Sigma_{m,n}^{-1} \varsigma_{x',z'}\right) ,
\]
where
\[
\Sigma_{m,n} = \begin{bmatrix}
K_{xx} + \eta \gamma I_n & H_{xz} \\ H_{xz}^\top & G_{zz} + \eta \rho W^{-1}
\end{bmatrix}, \quad \varsigma_{x',z'} = \begin{bmatrix}
k(x',X) & (\mathsf{Id} \otimes \mathcal{A}) k(x', Z) \\
(\mathsf{Id} \otimes \mathcal{A}) k(X, z')^\top & (\Acl \otimes \Acl) k(z', Z)
\end{bmatrix}.
\]

\emph{One of our main contributions of this work is to propose the negative log-marginal likelihood of this GP as an intrinsic uncertainty-aware measurement of model quality.} 
We refer to this as the \emph{Physics-Informed Log Evidence (PILE)}. Inspired by standard protocol for tuning GPs, the PILE score doubles as a model selection criterion for optimizing hyperparameters, including $\rho$, $\gamma$, $\eta$, and any other hyperparameters for the kernel $k$, including bandwidth.
\begin{definition}[Physics-Informed Log Evidence (PILE)] 
   The \emph{Physics-Informed Log Evidence (PILE) criterion} is $\frac{2}{m+n} \mathcal{F}_{m,n}$ where $\mathcal{F}_{m,n}$ is the Bayes free energy of the GP in \eqref{eq:GP}:
   \[
   \mathfrak{P}_{m,n} \coloneqq \frac{1}{m+n} \tilde{Y}^\top \Sigma_{m,n}^{-1} \tilde{Y} + \frac{1}{m+n} \log \det \Sigma_{m,n} + \log(2\pi \eta).
   \]
   where $\tilde{Y} = (y_1,\dots,y_n,r_1,\dots,r_m)^\top$. The PILE criterion is to be interpreted as \emph{lower is better}.
\end{definition}
Unlike the empirical risk (\ref{eq:ApproxLoss}), which can be computed in quadratic time by the equation (\ref{eqn:representer-theorem}), computing the PILE score generally requires cubic time. Fortunately, for large $m,n$, several algorithms exist to compute the marginal likelihood quickly and under memory constraints \citep{gardner2018gpytorch,ameli2025determinant}.

\ifconference
\vspace{-.2cm}
\else
\fi

\paragraph{Connection to the Fredholm Determinant.}
Let us now consider the special case where no data $(x_i,y_i)$ are prescribed and the PIKL framework is applied to find \emph{a} solution $f \in H$ to $\mathcal{A} f = 0$. 
In this ``data-free'' case, $r_i = 0$ can also be chosen and the PILE score simplifies to its last two terms. By taking $m \to \infty$, minimizers of (\ref{eq:ApproxLoss}) should become minimizers of our original problem (\ref{eq:MainLoss}). This is imposed by the following assumption.
\begin{assumption}
For any $f \in C^0(\bar{\Omega})$, as $m \to \infty$, the quadrature rule converges, that is,
$
\sum_{j=1}^m w_j f(z_j) \overset{m\to\infty}{\longrightarrow} 
\int_{\bar{\Omega}} f(z) \dd \mu(z)$.
\end{assumption}

Our next objective is to show that as $m \to \infty$, the PILE score in this special case converges in a suitable sense to a \emph{Fredholm determinant}, which provides a surprising quantifier of the effectiveness of a particular choice of kernel $k$ to solve a given problem.
The Fredholm determinant is a fascinating object with a complex history; and it typically appears only in the study of random matrices and determinantal point processes~\citep{DM21_NoticesAMS}. We refer to Appendix \ref{sec:FredholmApp} for its definition (and for the proof of the following result) and to \citet{bornemann2010numerical} for details of its computation in practice.

\begin{theorem}
\label{thm:Fredholm}
Let $\mathcal{G}:L^2(\bar{\Omega},\mu,\mathbb{R}^{p+1}) \to L^2(\bar{\Omega},\mu,\mathbb{R}^{p+1})$ be the integral operator
\[
(\mathcal{G}f)(x, z) = \frac{1}{\eta\rho} \int_{\Omega \times \boldsymbol{\Gamma}} (\mathcal{A} \otimes \mathcal{A}) k(z, z') f(z') \dd z'.
\]
Letting $C_m = m\eta\rho-\sum_{i=1}^{m}\log w_{i}+m\log(2\pi\eta)$, as $m \to \infty$, 
\vspace{.1cm}
the sequence of normalized PILE scores converge to the \textbf{Fredholm determinant}: $m\mathfrak{P}_{m,0} - C_m \overset{m\to\infty}{\longrightarrow} \mathfrak{P}_0 = \log \det(I + \mathcal{G})$.
\end{theorem}
\ifconference
\else
\vspace{4mm}
\fi
Note that one can also choose $\rho=\frac{1}{m\eta}\sum_{i=1}^{m}\log(\frac{w_{i}}{2\pi\eta})$ so that the normalized PILE score becomes equal to the PILE score. 
This choice of $\rho$ provides an avenue to calibrate uncertainty. 
Theorem \ref{thm:Fredholm} provides a new interpretation of the Fredholm determinant of integro-differential operators of the form of $\mathcal{G}$ in terms of the base model difficulty of solving differential equations within the corresponding RKHS induced by $k$. 
In turn, model selection for a given problem prescribed by the operator $\mathcal{A}$ can be achieved in a \emph{data-independent} fashion by comparing values of $\mathfrak{P}_0$. 
In Section \ref{app:case-study}, we provide a case study, in which we use the Fredholm determinant to select the best kernel from a family of \textit{anisotropic RBF kernels}. When used to solve a 1D convection PDE, we observe a drastic improvement in the capacity of the PIML model to fit the target solution.

\ifconference
\vspace{-.25cm}
\else
\fi

\section{Case Studies}
\label{sec:empirical}
\ifconference
\vspace{-.25cm}
\else
\fi

In this section, we present several case studies to illustrate the utility of the PILE score.
We demonstrate how the PILE score can be used to select hyperparameters such as kernel bandwidth, loss regularization, and can even be used to select the kernel function from a parametrized family.




\ifconference
\vspace{-.2cm}
\else
\fi

\subsection{Hyperparameter Selection using the PILE Score}
\ifconference
\vspace{-.2cm}
\else
\fi

%
%

%

\begin{figure}[t]
    \centering
    \includegraphics[width=\linewidth]{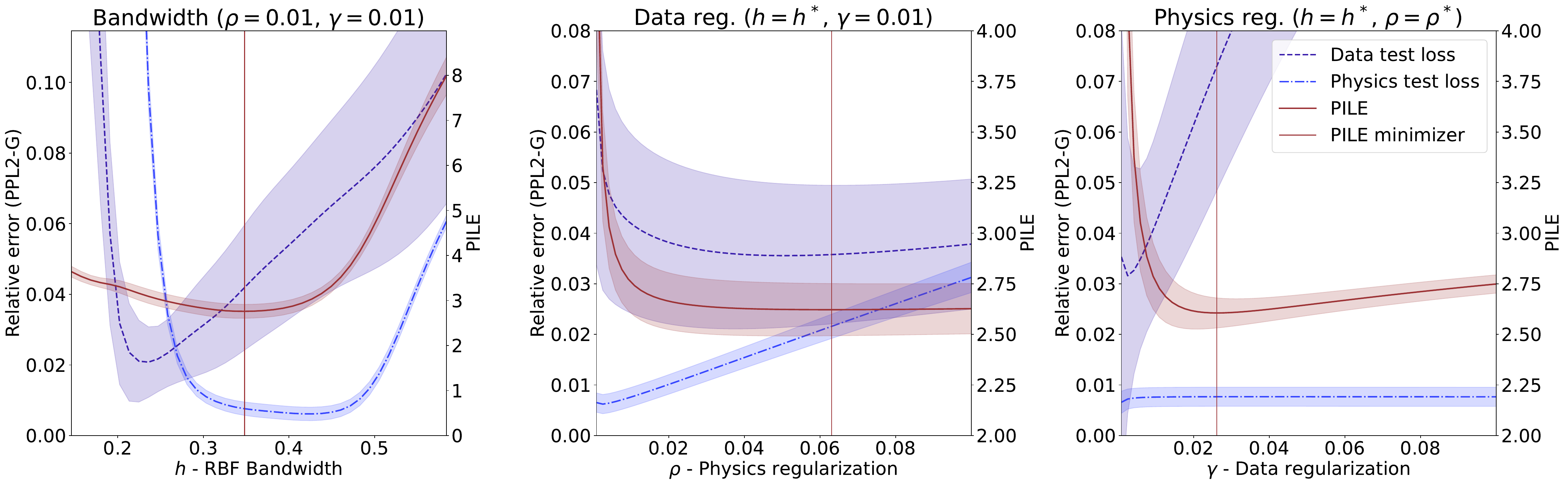}
    \caption{\textbf{Automatic hyperparameter selection with PILE.} The PILE score and relative PPL2-G error sources for varying bandwidth, physics regularization, and data regularization parameters. Error bars show $\pm 2\hat{\sigma}$ coverage, where $\hat{\sigma}$ is the empirical standard deviation of PPL2-G (blue bars) and PILE (red bars). \textbf{(Left)} Bandwidth selection via minimizing the PILE score provides an accurate fit, balancing the data and physics generalization errors. \textbf{(Middle, Right)} After selecting the optimal bandwidth $h^*$, we sequentially minimize PILE first with respect to the physics regularization parameter $\rho$, and then with respect to the data regularization parameter $\gamma$. For small values of $\rho$ and $\gamma$, PILE diverges as the regression model overfits the noisy observations.}
\label{fig:pile-guided-hyperparameters-poisson}
\end{figure}

Our first case study examines the baseline efficacy of the PILE score, and the Bayes free energy at large, for solving the multi-objective ambiguity in the PIML problem. 
See Figure~\ref{fig:pile-guided-hyperparameters-poisson}.
The point of this case study is to demonstrate the practicality of PILE by using it to automatically select all the relevant hyperparameters for the problem. 
As a simple test case, consider a Poisson equation with Dirichlet boundary conditions,
on the domain $\Omega = (-1, 1)^{2}$,
\begin{equation}
\label{eqn:poiss}
\Delta f(x) = g(x)\text{ for } x \in \Omega,\qquad\qquad f(x) = 0\text{ for } x \in \partial \Omega,
\end{equation}
and with forcing function $g(x) = 10 + 10 \sin (2 \pi x) \sin (2 \pi y)$. 
A 2D (type-1) Chebyshev quadrature scheme is used to determine the gridpoints $z_{(i,j)}$ and corresponding weights, $w_{(i,j)}$ for $i,j=1,\dots,m$, at which to evaluate the derivative loss:
letting $s_i = \cos(\frac{2k+1}{2 m_{\text{grid}}} \pi)$, we set $z_{(i,j)} = (s_i,s_j) \in \R^2$ and $w_{(i,j)} = 4 / m^2$. 
The type-1 Chebyshev grid points are known to have excellent properties for numerical approximation of integrals on $(-1, 1)$: for example,
if $f, g \in C^\infty(\Omega)$, then the integration error $\|g - \mathcal{D} f\|_{L^2(\Omega)}$ converges as $e^{-\Omega(m)}$ \citep[Chapters 7, 8]{trefethen2019approximation}. 
This choice of quadrature scheme enables us to closely approximate the true $L^2(\bar{\Omega})$ norm of the PDE residual. 

In the physics-informed setting, there are two sources of error: \textit{data error}, measuring the fit of $\hat{f}$ to $f$; and \textit{physics error}, measuring the fit of $\hat{g}$ to $g =\Dcl f$. 
Following \citet{pmlr-v202-hodgkinson23a}, we define the (unnormalized) \textit{data PPL2-G} error as
\[
\tilde{\mathcal{R}}_{\text{data}}(\hat{f}) \coloneqq  \textstyle\int_{\bar{\Omega}}\mathbb{E}_{\hat{f}(z)}[(\hat{f}(z) - f(z))^2] \, dz
    \approx \textstyle\sum_{i,j=1}^{m_{\text{eval}}} w_i \mathbb{E}[(\hat{f}(z_{(i,j)})-f(z_{(i,j)})^2 ],\]
over $\hat{f}(z) \sim \Ncl(m^{\mid y, r}, \Sigma^{\mid y, r})$, marginalized over $\hat{g}$. Analogously,
\[
\tilde{\mathcal{R}}_{\text{phys}}(\hat{g}) \coloneqq
   \textstyle\int_{\bar{\Omega}}\mathbb{E}_{\hat{g}(z)}\left[(\hat{g}(z) - \Dcl f(z))^2 \right] \, dx 
    \approx \textstyle\sum_{i,j=1}^{m_{\text{eval}}} w_i \mathbb{E}[(\hat{f}(z_{(i,j)})-f(z_{(i,j)})^2 ],
    \]
is the (unnormalized) \textit{physics PPL2-G} error, marginalized over $\hat{f}$. The quadrature points $z_{(i,j)}$ depend on $m_{\text{eval}}$, chosen to be large ($m_{\text{eval}} = 30 \gg m)$ to ensure accurate $L^2(\bar{\Omega})$ approximations. 
For appropriate comparison, we normalize errors: $\mathcal{R}_{\text{data}}(\hat{f})  = \frac{\tilde{\mathcal{R}}_{\text{data}}(\hat{f})}{\|f\|_{L^2(\bar{\Omega})}}$, and $\mathcal{R}_{\text{phys}}(\hat{f})  = \frac{\tilde{\mathcal{R}}_{\text{phys}}(\hat{f})}{\|\Dcl f\|_{L^2(\bar{\Omega})}}$. 
Both should be small, but since they cannot generally vanish simultaneously, \emph{the optimal point on the Pareto front of these losses (the best solution to the PDE) is unclear}. 
Figure \ref{fig:pile-guided-hyperparameters-poisson} demonstrates how PILE addresses this issue.

\begin{figure}
    \centering
    \ifconference
    \includegraphics[width=0.8\linewidth]{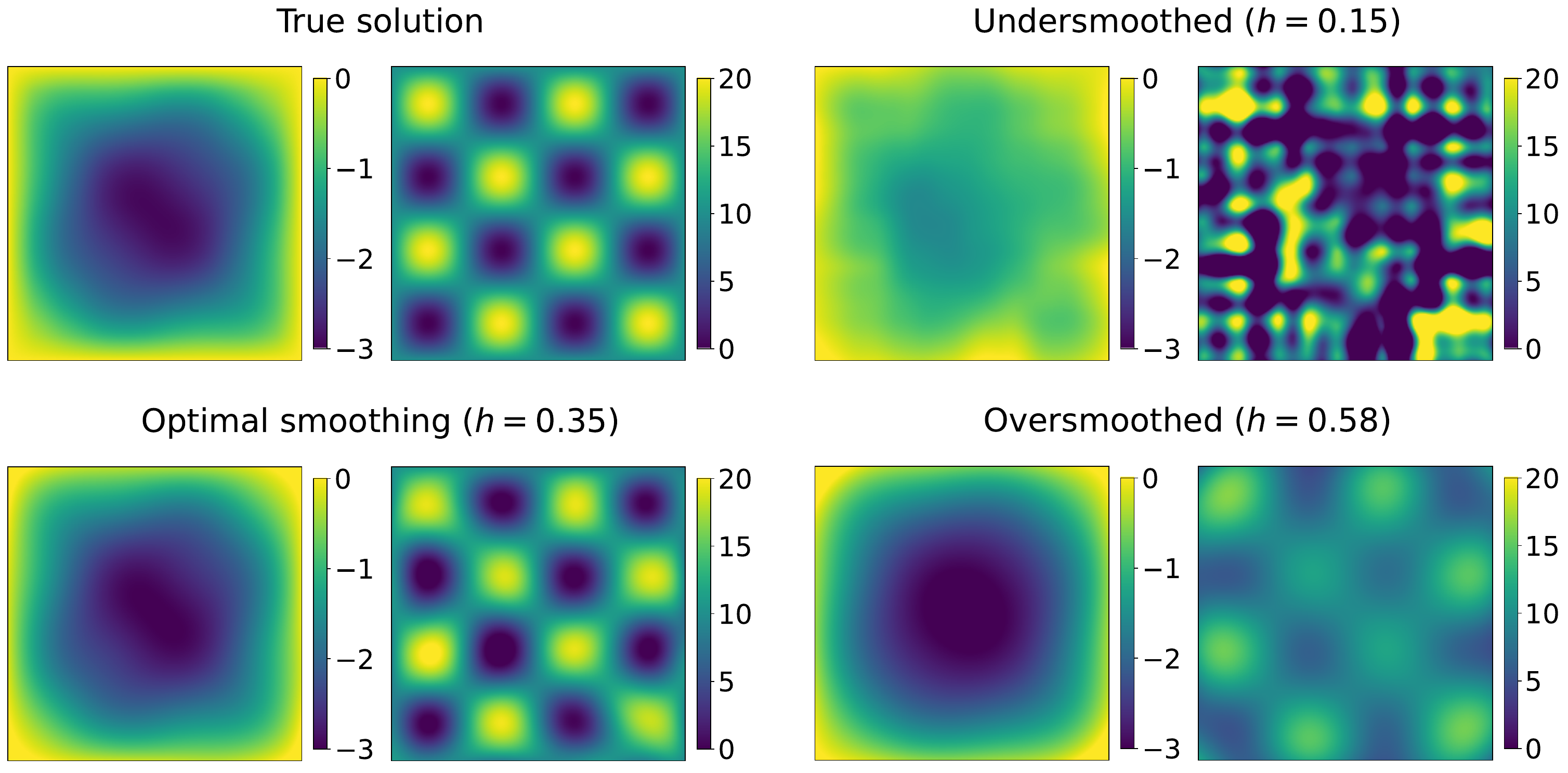}
\else
\includegraphics[width=\linewidth]{figures/Smoothing_effects.pdf}
\fi
    \caption{\textbf{Optimizing PILE prevents under- and over-smoothing.} Qualitative plot of the negative effects of oversmoothing and undersmoothing in PIKL. Each panel shows $\hat{f}$, $\hat{g}$ on the left and right, respectively. When $h$ is too small, the derivative estimate is undersmoothed and irregular. When $h$ is too large, oversmoothing effects prevent the model from fitting the derivative.  The optimal value of $h$ is identified by optimizing PILE.
    \ifconference
\vspace{-.5cm}
\else
\fi}
    \label{fig:pile-guided-hyperparameters}
\end{figure}

This case study also allows us to consider a related hyperparameter challenge in RKHSs and GP regression, namely selecting the \textit{bandwidth} of a shift invariant kernel. 
Recall that a shift invariant kernel with bandwidth $h > 0$ has the form $k_h(x, y) = k(\frac{x-y}{h})$. 
These kernels enjoy special analytical properties and fast randomized approximations \citep{rahimi2007random}, making them popular in practice. 
As with the hyperparameter selection problem described above, here too it is important to tune the kernel bandwidth for optimal performance: if it is too large, the kernel becomes \textit{oversmoothed}, and the resulting regression estimator has high bias; whereas if it is too small, the resulting regression estimator may overfit observations or take near-zero values on unseen data. 
To illustrate how PILE performs in this context, see Figure~\ref{fig:pile-guided-hyperparameters}.
We train the regression model using grid size $n=m=13$, so that there are a total of $13^2$ noisy observations $y_{(i,j)} = f(z_{(i,j)}) + \eps_{(i,j)}$, $z_{(i,j)} = \Dcl f(z_{(i,j)}) + \eps_{(i,j)}'$ for independent $\eps_{(i,j)}, \eps_{(i,j)}' \sim \Ncl(0, 1)$, with $i,j = 1, \ldots, 13$. 
Despite the high noise level and the relatively small number of samples, the regularized regression estimator matches the target function when fit with the optimal bandwidth according to the PILE score, avoiding the undersmoothing as well as the oversmoothing regime.
That is, the PILE score has successfully overcome the multi-objective problem. 
More generally, when used to select $\rho, \gamma > 0$, divergence of the PILE score is an accurate indicator of model overfitting.  

\ifconference
\begin{wrapfigure}{r}{0.4\textwidth}
    \centering\vspace{-.2cm}
    \includegraphics[width=0.35\textwidth]{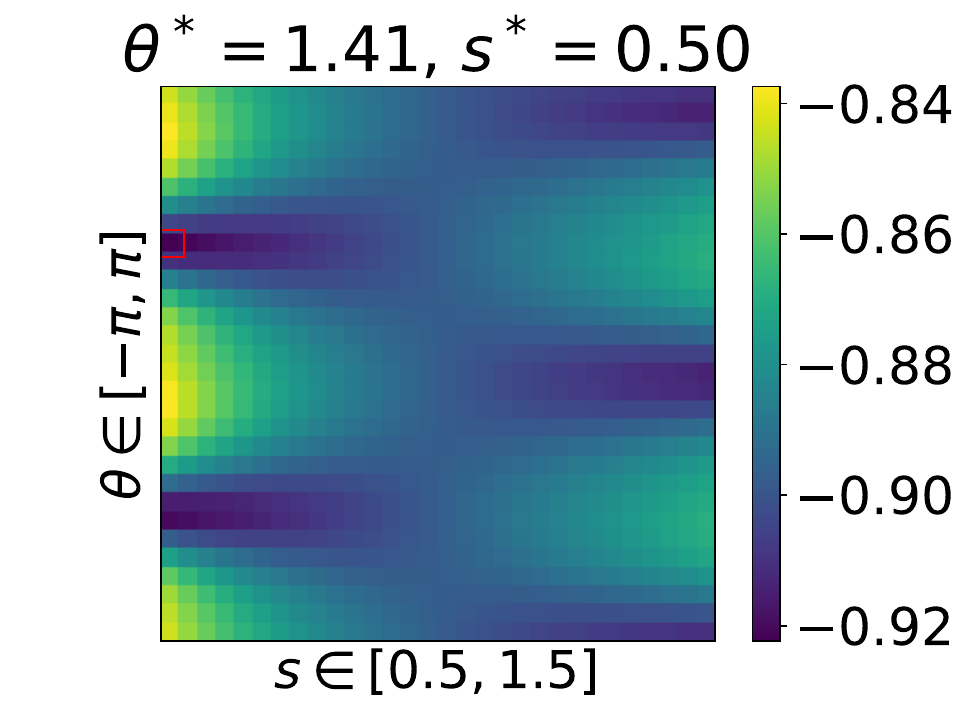}\vspace{-.2cm}
    \caption{\textbf{Data-free PILE landscape for anisotropic RBF kernel}. 
    Fredholm determinants of $k_{\theta, s}$ plotted for $\theta \in [-\pi, \pi]$ and $s \in [0.5, 1.5]$. 
    This quantity is empirically minimized at $\theta^* \approx 1.41$, $s^* \approx 0.5$ (shown in red), but there are evidently symmetries in the landscape.
    \vspace{-1cm}}
    \label{fig:fredholm-determinant}
\end{wrapfigure}
\else
\begin{figure}[h]
    \centering
    \includegraphics[width=0.4\textwidth]{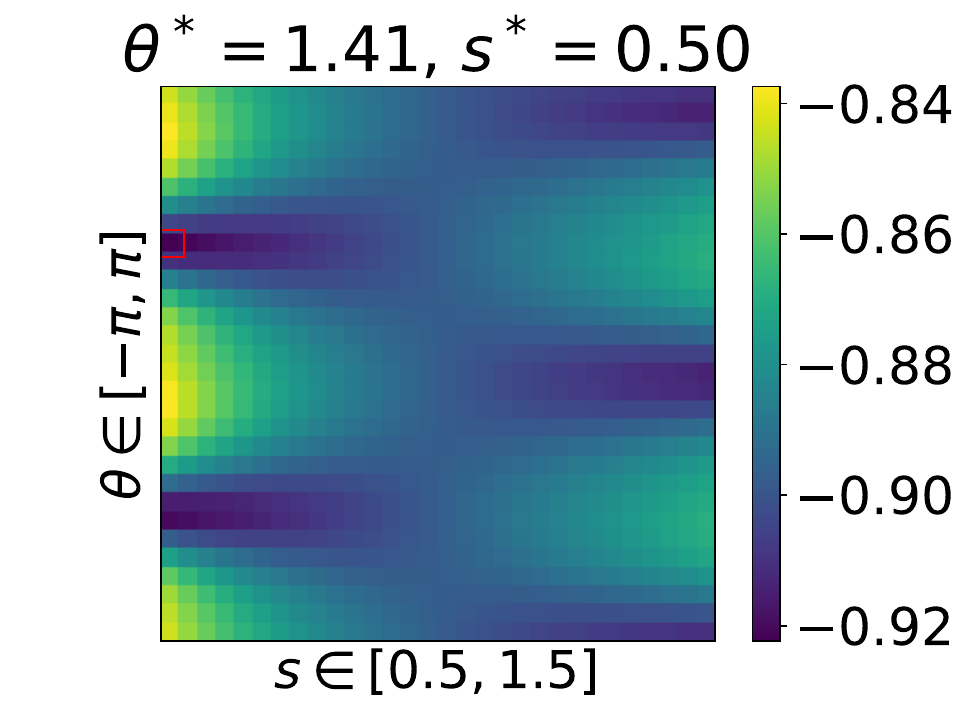}
    \caption{\textbf{Data-free PILE landscape for anisotropic RBF kernel}. 
    Fredholm determinants of $k_{\theta, s}$ plotted for $\theta \in [-\pi, \pi]$ and $s \in [0.5, 1.5]$. 
    This quantity is empirically minimized at $\theta^* \approx 1.41$, $s^* \approx 0.5$ (shown in red), but there are evidently symmetries in the landscape.}
    \label{fig:fredholm-determinant}
\end{figure}
\fi


\ifconference
\vspace{-.2cm}
\else
\fi

\subsection{Diagnosing and avoiding model failure with the data-free PILE score}
\ifconference
\vspace{-.1cm}
\else
\fi

\label{app:case-study}

Our second case study analyzes the well-known wave equation baseline introduced by \citet{krishnapriyan2021characterizing}, which was shown to ``break'' vanilla multilayer perceptron methods, within a PINN context. 
We observe that for an isotropic RBF kernel, there is no bandwidth that simultaneously achieves good physics and data fits, and PILE diagnoses this by selecting an oversmoothed ``all-zero'' solution. 
Perhaps more surprisingly, when we consider a broader class of ``anisotropic RBF'' kernels, we find that optimizing the data-free PILE score (Figure \ref{fig:fredholm-determinant}) with respect to the kernel function yields a model with excellent physics and data fit. 
The PILE-optimal kernel can be identified automatically, \emph{prior to data acquisition, and with no domain knowledge.} 

Following \cite[Section 3.1]{krishnapriyan2021characterizing}, consider the convection PDE of the form
\begin{align}\label{eqn:convection-eqn}
    \begin{cases}
        \frac{\partial f}{\partial t} (t, x) + \beta \frac{\partial f}{\partial x}(t, x) = 0, & t \in [0,1],\ x \in [0, 2 \pi], \\ 
        f(0, x) = \sin(x). 
    \end{cases}
\end{align}
\ifconference
\vspace{-.5cm}
\else
\fi

\begin{figure}[h]
    \centering
    \includegraphics[width=\linewidth]{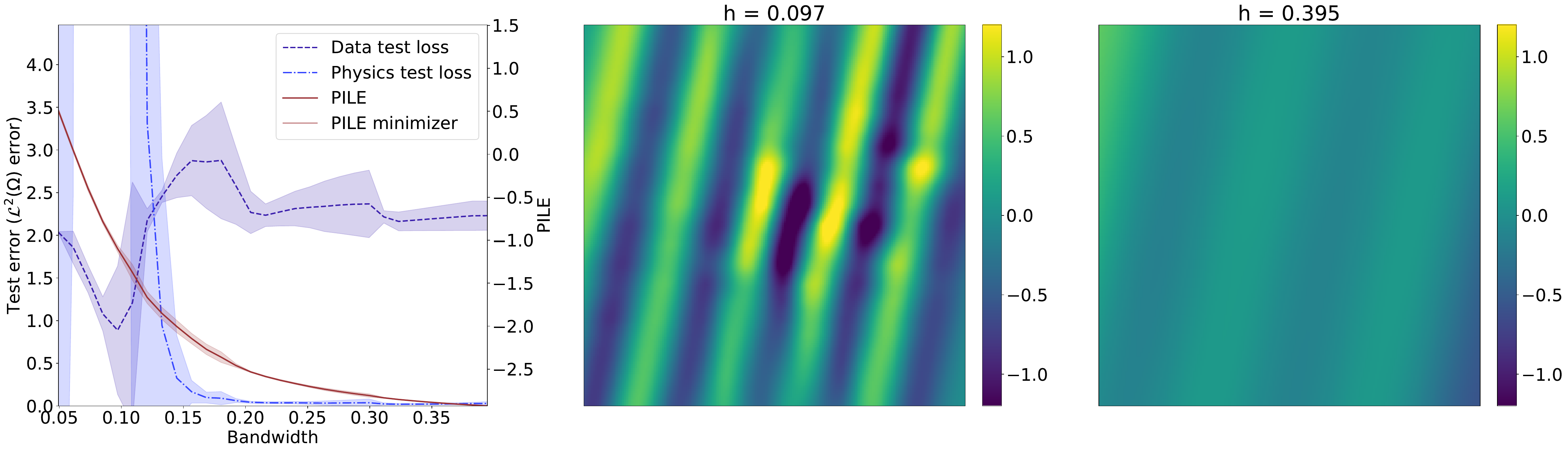}
    \caption{\textbf{Diagnostics with PILE: identifying model failure.} Fitting the convection PDE, \eqref{eqn:convection-eqn}, with an RBF kernel. There is no appropriate bandwidth for this problem; and the PILE score diagnosis this by selecting the ``all zeros'' solution.
    }
    \label{fig:pathological-rbf-convection}
\end{figure}

We set $n = 1000$, $m=20^2$, and we assume access to observations of the form $y_i = f(x_i) + \eps_i$, $i=1\ldots n$, for $x_i \sim \text{Unif}([0,1] \times [0,2\pi])$. 
We observe empirically that fitting an RBF kernel 
leads to pathological behavior.
This is summarized in Figure \ref{fig:pathological-rbf-convection}. 
As shown in Figure \ref{fig:pathological-rbf-convection} (left), the data loss is only small at bandwidths $h \approx 0.1$, while physics loss blows up at bandwidths $\leq h \approx 0.15$.

\begin{figure}[h]
    \centering
    \ifconference
\includegraphics[width=0.6\linewidth]{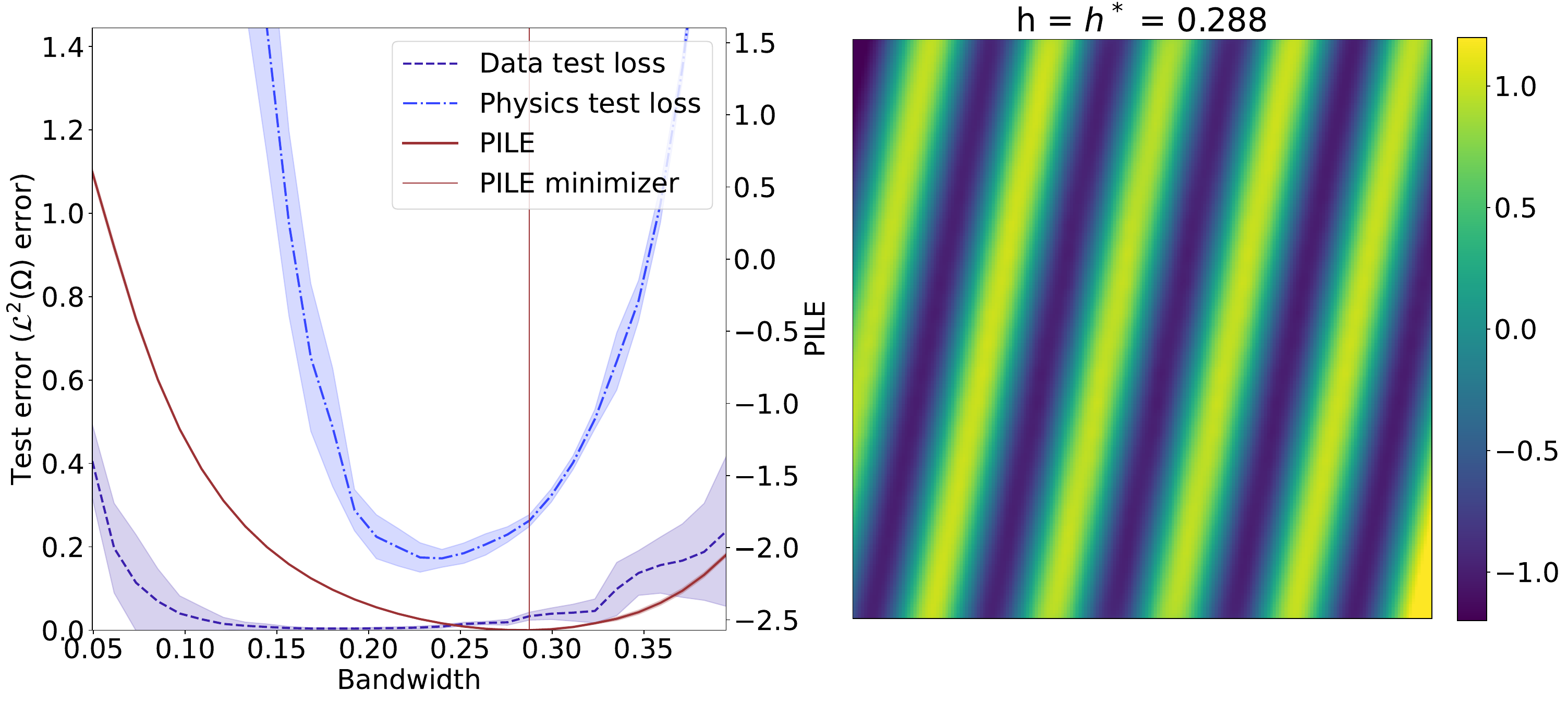}
\else
\includegraphics[width=0.67\linewidth]{figures/aniso-rbf-convection.pdf}
\fi
    \caption{\textbf{Diagnostics with PILE: hyperparameter selection after kernel adjustment.} Fitting the convection PDE, \eqref{eqn:convection-eqn}, with an anisotropic RBF kernel. By choosing the kernel with minimum Fredholm determinant, i.e., the ``data-free'' variant of the PILE score, we can automatically identify a ``good'' kernel for the continuity~PDE.
    \ifconference
\vspace{-.3cm}
\else
\fi}
    \label{fig:aniso-rbf}
\end{figure}

On the other hand, we could consider an \textit{anisotropic} family of kernels defined by hyperparameters $\theta \in [-\pi, \pi]$, $s > 0$, and given by 
\begin{align*}
    k_{\theta, s}(x,y) \coloneqq e^{-\frac{1}{2}(x-y)^T \Sigma_{\theta, s}(x-y)}, \quad \Sigma_{\theta, s} \coloneqq \begin{bmatrix}
        \cos(\theta) & - \sin(\theta) \\ 
        \sin(\theta) & \cos(\theta)
    \end{bmatrix} \begin{bmatrix}
        s^2 & 0 \\ 0 & s^{-2}
    \end{bmatrix}\begin{bmatrix}
        \cos(\theta) & - \sin(\theta) \\ 
        \sin(\theta) & \cos(\theta)
    \end{bmatrix}.
\end{align*}
To find an appropriate kernel among this family, we select $\theta, s$ to minimize the data-free PILE score, whose loss landscape is shown in Figure \ref{fig:fredholm-determinant}. 
After kernel selection, the loss basins of the physics and data loss become drastically better conditioned, leading to an excellent model fit.
This is shown in Figure \ref{fig:aniso-rbf}.

\section{Conclusions}
\ifconference
\vspace{-.2cm}
\else
\fi

We have introduced the PILE score, a model selection metric and diagnostic tool for PIKL. 
Our method amounts to hyperparameter selection via Empirical Bayes, thereby removing ambiguities involved in parameter selection under two competing (and qualitatively-different) loss terms. 
While already practical, it is useful to recognize how PILE can extend to other contexts. 

\ifconference
\begin{itemize}[leftmargin=*,noitemsep,nolistsep]
\else
\begin{itemize}
[leftmargin=*]
\fi
    \item 
    \textbf{Nonlinear Operators.} 
    We have studied linear differential operators with kernel-based models, but PIKL also extends to nonlinear PDEs \citep{chen2021solving}, following the framework of \citet{dashti2013map}. 
    The nonlinear operator $\Acl$ can be linearized via its Fr\`{e}chet derivative, $D\Acl$, at the solution $\hat{f}$; and the PIKL setup and PILE score can be applied using $D\Acl$ in place of $\Acl$. The justification for this PILE score follows the arguments of \citet{wacker2017laplace}. 
    \vspace{.1cm}
    \item
    \textbf{Neural Networks.} 
    The PIKL framework is distinct from both PINNs and Neural ODEs (since neural network regression differs from KRR). Approximations of the free energy are typically derived using Laplace approximations, such as in the derivation of the BIC \citep{schwarz1978estimating}, although extensions exist when these approximations fail \citep{drton2017bayesian}. Of particular relevance is the Interpolating Information Criterion (IIC) \citep{hodgkinson2023interpolating}, designed for models that interpolate data, akin to GP-based PIML. Inspired by IIC, one could evaluate the PILE score for trained PINNs using the empirical neural tangent kernel \citep{novak2022fast,jacot2018neural} in place of $k$. At present, this is expensive, although approximations such as those seen in \citet{ameli2025determinant} suggests this might yet become tractable, under very reasonable assumptions.
\end{itemize}


%
%

\paragraph{Acknowledgements.}
MD was partially supported by a DOE Computational Science Graduate Fellowship.
MWM would like to acknowledge the DOD DARPA AIQ program, the DOE LBNL LDRD program, and the NSF for partial support of this work.


\bibliography{main}

\begin{thebibliography}{52}
\providecommand{\natexlab}[1]{#1}
\providecommand{\url}[1]{\texttt{#1}}
\expandafter\ifx\csname urlstyle\endcsname\relax
  \providecommand{\doi}[1]{doi: #1}\else
  \providecommand{\doi}{doi: \begingroup \urlstyle{rm}\Url}\fi

\bibitem[Ameli et~al.(2025)Ameli, van~der Heide, Hodgkinson, Roosta, and Mahoney]{ameli2025determinant}
Siavash Ameli, Chris van~der Heide, Liam Hodgkinson, Fred Roosta, and Michael~W Mahoney.
\newblock Determinant estimation under memory constraints and neural scaling laws.
\newblock \emph{arXiv preprint arXiv:2503.04424}, 2025.

\bibitem[Benitez et~al.(2025)Benitez, Guo, Hegazy, Dokmanic, Mahoney, and de~Hoop]{neurde_TR}
J.~A.~L. Benitez, J.~Guo, K.~Hegazy, I.~Dokmanic, M.~W. Mahoney, and M.~V. de~Hoop.
\newblock Neural equilibria for long-term prediction of nonlinear conservation laws.
\newblock Technical Report Preprint: arXiv:2501.06933, 2025.

\bibitem[Bogachev(1998)]{bogachev1998gaussian}
V.I. Bogachev.
\newblock \emph{Gaussian Measures}.
\newblock Mathematical surveys and monographs. American Mathematical Society, 1998.
\newblock ISBN 9780821810545.
\newblock URL \url{https://books.google.com/books?id=otmkhedD8ZAC}.

\bibitem[Bornemann(2010)]{bornemann2010numerical}
Folkmar Bornemann.
\newblock On the numerical evaluation of {F}redholm determinants.
\newblock \emph{Mathematics of Computation}, 79\penalty0 (270):\penalty0 871--915, 2010.

\bibitem[Chen et~al.(2018)Chen, Rubanova, Bettencourt, and Duvenaud]{chen2018neural}
Ricky~TQ Chen, Yulia Rubanova, Jesse Bettencourt, and David~K Duvenaud.
\newblock Neural ordinary differential equations.
\newblock \emph{Advances in neural information processing systems}, 31, 2018.

\bibitem[Chen et~al.(2021)Chen, Hosseini, Owhadi, and Stuart]{chen2021solving}
Yifan Chen, Bamdad Hosseini, Houman Owhadi, and Andrew~M Stuart.
\newblock {Solving and learning nonlinear {PDEs} with {G}aussian processes}.
\newblock \emph{Journal of Computational Physics}, 447:\penalty0 110668, 2021.

\bibitem[Dashti et~al.(2013)Dashti, Law, Stuart, and Voss]{dashti2013map}
Masoumeh Dashti, Kody~JH Law, Andrew~M Stuart, and Jochen Voss.
\newblock {MAP estimators and their consistency in Bayesian nonparametric inverse problems}.
\newblock \emph{Inverse Problems}, 29\penalty0 (9):\penalty0 095017, 2013.

\bibitem[Derezinski \& Mahoney(2021)Derezinski and Mahoney]{DM21_NoticesAMS}
M.~Derezinski and M.~W. Mahoney.
\newblock Determinantal point processes in randomized numerical linear algebra.
\newblock \emph{Notices of the AMS}, 68\penalty0 (1):\penalty0 34--45, 2021.

\bibitem[Doum{\`e}che et~al.(2024{\natexlab{a}})Doum{\`e}che, Bach, Biau, and Boyer]{doumeche2024physics}
Nathan Doum{\`e}che, Francis Bach, G{\'e}rard Biau, and Claire Boyer.
\newblock Physics-informed kernel learning.
\newblock \emph{arXiv preprint arXiv:2409.13786}, 2024{\natexlab{a}}.

\bibitem[Doum{\`e}che et~al.(2024{\natexlab{b}})Doum{\`e}che, Bach, Biau, and Boyer]{pmlr-v247-doumeche24a}
Nathan Doum{\`e}che, Francis Bach, G{\'e}rard Biau, and Claire Boyer.
\newblock Physics-informed machine learning as a kernel method.
\newblock In \emph{Proceedings of Thirty Seventh Conference on Learning Theory}, volume 247 of \emph{Proceedings of Machine Learning Research}, pp.\  1399--1450. PMLR, 2024{\natexlab{b}}.

\bibitem[Drton \& Plummer(2017)Drton and Plummer]{drton2017bayesian}
Mathias Drton and Martyn Plummer.
\newblock A {B}ayesian information criterion for singular models.
\newblock \emph{Journal of the Royal Statistical Society Series B: Statistical Methodology}, 79\penalty0 (2):\penalty0 323--380, 2017.

\bibitem[Efron(2024)]{efron2024empirical}
Bradley Efron.
\newblock Empirical {B}ayes: Concepts and methods.
\newblock In \emph{Handbook of Bayesian, Fiducial, and Frequentist Inference}, pp.\  8--34. Chapman and Hall/CRC, 2024.

\bibitem[Gardner et~al.(2018)Gardner, Pleiss, Weinberger, Bindel, and Wilson]{gardner2018gpytorch}
Jacob Gardner, Geoff Pleiss, Kilian~Q Weinberger, David Bindel, and Andrew~G Wilson.
\newblock {GPyTorch: Blackbox matrix-matrix Gaussian process inference with GPU acceleration}.
\newblock \emph{Advances in neural information processing systems}, 31, 2018.

\bibitem[Garnelo et~al.(2018)Garnelo, Schwarz, Rosenbaum, Viola, Rezende, Eslami, and Teh]{garnelo2018neural}
Marta Garnelo, Jonathan Schwarz, Dan Rosenbaum, Fabio Viola, Danilo~J Rezende, SM~Eslami, and Yee~Whye Teh.
\newblock Neural processes.
\newblock \emph{arXiv preprint arXiv:1807.01622}, 2018.

\bibitem[Gelman et~al.(1995)Gelman, Carlin, Stern, and Rubin]{gelman1995bayesian}
Andrew Gelman, John~B Carlin, Hal~S Stern, and Donald~B Rubin.
\newblock \emph{Bayesian data analysis}.
\newblock Chapman and Hall/CRC, 1995.

\bibitem[Geneva \& Zabaras(2020)Geneva and Zabaras]{geneva2020modeling}
Nicholas Geneva and Nicholas Zabaras.
\newblock Modeling the dynamics of {PDE} systems with physics-constrained deep auto-regressive networks.
\newblock \emph{Journal of Computational Physics}, 403:\penalty0 109056, 2020.

\bibitem[Gribov \& Krivoruchko(2020)Gribov and Krivoruchko]{gribov2020empirical}
Alexander Gribov and Konstantin Krivoruchko.
\newblock Empirical {B}ayesian kriging implementation and usage.
\newblock \emph{Science of the Total Environment}, 722:\penalty0 137290, 2020.

\bibitem[Hansen et~al.(2023)Hansen, Maddix, Alizadeh, Gupta, and Mahoney]{hansen2023learning}
Derek Hansen, Danielle~C Maddix, Shima Alizadeh, Gaurav Gupta, and Michael~W Mahoney.
\newblock Learning physical models that can respect conservation laws.
\newblock In \emph{International Conference on Machine Learning}, pp.\  12469--12510. PMLR, 2023.

\bibitem[H\"{a}rk\"{o}nen et~al.(2023)H\"{a}rk\"{o}nen, Lange-Hegermann, and Rait\u{a}]{harkonen2023gaussian}
Marc H\"{a}rk\"{o}nen, Markus Lange-Hegermann, and Bogdan Rait\u{a}.
\newblock Gaussian process priors for systems of linear partial differential equations with constant coefficients.
\newblock In \emph{Proceedings of the 40th International Conference on Machine Learning}, ICML'23. JMLR.org, 2023.

\bibitem[Hodgkinson et~al.(2023{\natexlab{a}})Hodgkinson, Van Der~Heide, Roosta, and Mahoney]{pmlr-v202-hodgkinson23a}
Liam Hodgkinson, Chris Van Der~Heide, Fred Roosta, and Michael~W. Mahoney.
\newblock Monotonicity and double descent in uncertainty estimation with {G}aussian processes.
\newblock In \emph{Proceedings of the 40th International Conference on Machine Learning}, volume 202 of \emph{Proceedings of Machine Learning Research}, pp.\  13085--13117. PMLR, 2023{\natexlab{a}}.

\bibitem[Hodgkinson et~al.(2023{\natexlab{b}})Hodgkinson, van~der Heide, Salomone, Roosta, and Mahoney]{hodgkinson2023interpolating}
Liam Hodgkinson, Chris van~der Heide, Robert Salomone, Fred Roosta, and Michael~W Mahoney.
\newblock The interpolating information criterion for overparameterized models.
\newblock \emph{arXiv preprint arXiv:2307.07785}, 2023{\natexlab{b}}.

\bibitem[Jacot et~al.(2018)Jacot, Gabriel, and Hongler]{jacot2018neural}
Arthur Jacot, Franck Gabriel, and Cl{\'e}ment Hongler.
\newblock Neural tangent kernel: Convergence and generalization in neural networks.
\newblock \emph{Advances in neural information processing systems}, 31, 2018.

\bibitem[Jin et~al.(2022)Jin, Banerjee, and Montufar]{jinlearning}
Hui Jin, Pradeep~Kr Banerjee, and Guido Montufar.
\newblock {Learning Curves for {G}aussian Process Regression with Power-Law Priors and Targets}.
\newblock In \emph{International Conference on Learning Representations}, 2022.

\bibitem[Jin et~al.(2021)Jin, Cai, Li, and Karniadakis]{jin2021nsfnets}
Xiaowei Jin, Shengze Cai, Hui Li, and George~Em Karniadakis.
\newblock {NSFnets (Navier-Stokes flow nets): Physics-informed neural networks for the incompressible Navier-Stokes equations}.
\newblock \emph{Journal of Computational Physics}, 426:\penalty0 109951, 2021.

\bibitem[Kanagawa et~al.(2018)Kanagawa, Hennig, Sejdinovic, and Sriperumbudur]{kanagawa2018gaussian}
Motonobu Kanagawa, Philipp Hennig, Dino Sejdinovic, and Bharath~K Sriperumbudur.
\newblock Gaussian processes and kernel methods: A review on connections and equivalences.
\newblock \emph{arXiv preprint arXiv:1807.02582}, 2018.

\bibitem[Karniadakis et~al.(2021)Karniadakis, Kevrekidis, Lu, Perdikaris, Wang, and Yang]{karniadakis2021physics}
George~Em Karniadakis, Ioannis~G Kevrekidis, Lu~Lu, Paris Perdikaris, Sifan Wang, and Liu Yang.
\newblock Physics-informed machine learning.
\newblock \emph{Nature Reviews Physics}, 3\penalty0 (6):\penalty0 422--440, 2021.

\bibitem[Kim et~al.(2019)Kim, Mnih, Schwarz, Garnelo, Eslami, Rosenbaum, Vinyals, and Teh]{kim2019attentive}
Hyunjik Kim, Andriy Mnih, Jonathan Schwarz, Marta Garnelo, Ali Eslami, Dan Rosenbaum, Oriol Vinyals, and Yee~Whye Teh.
\newblock Attentive neural processes.
\newblock \emph{arXiv preprint arXiv:1901.05761}, 2019.

\bibitem[Knopp(2013)]{knopp2013theorie}
Konrad Knopp.
\newblock \emph{{Theorie und Anwendung der unendlichen Reihen}}, volume~2.
\newblock Springer-Verlag, 2013.

\bibitem[Kovachki et~al.(2023)Kovachki, Li, Liu, Azizzadenesheli, Bhattacharya, Stuart, and Anandkumar]{kovachki2023neural}
Nikola Kovachki, Zongyi Li, Burigede Liu, Kamyar Azizzadenesheli, Kaushik Bhattacharya, Andrew Stuart, and Anima Anandkumar.
\newblock Neural operator: Learning maps between function spaces with applications to {PDEs}.
\newblock \emph{Journal of Machine Learning Research}, 24\penalty0 (89):\penalty0 1--97, 2023.

\bibitem[Krishnapriyan et~al.(2021)Krishnapriyan, Gholami, Zhe, Kirby, and Mahoney]{krishnapriyan2021characterizing}
Aditi~S. Krishnapriyan, Amir Gholami, Shandian Zhe, Robert Kirby, and Michael~W Mahoney.
\newblock Characterizing possible failure modes in physics-informed neural networks.
\newblock \emph{Advances in Neural Information Processing Systems}, 34, 2021.

\bibitem[Krishnapriyan et~al.(2023)Krishnapriyan, Queiruga, Erichson, and Mahoney]{krishnapriyan2023learning}
Aditi~S Krishnapriyan, Alejandro~F Queiruga, N~Benjamin Erichson, and Michael~W Mahoney.
\newblock Learning continuous models for continuous physics.
\newblock \emph{Communications Physics}, 6\penalty0 (1):\penalty0 319, 2023.

\bibitem[Krivoruchko \& Gribov(2019)Krivoruchko and Gribov]{krivoruchko2019evaluation}
Konstantin Krivoruchko and Alexander Gribov.
\newblock Evaluation of empirical {B}ayesian kriging.
\newblock \emph{Spatial Statistics}, 32:\penalty0 100368, 2019.
\newblock ISSN 2211-6753.

\bibitem[Lotfi et~al.(2022)Lotfi, Izmailov, Benton, Goldblum, and Wilson]{lotfi2022bayesian}
Sanae Lotfi, Pavel Izmailov, Gregory Benton, Micah Goldblum, and Andrew~Gordon Wilson.
\newblock Bayesian model selection, the marginal likelihood, and generalization.
\newblock In \emph{International Conference on Machine Learning}, pp.\  14223--14247. PMLR, 2022.

\bibitem[Lu et~al.(2021)Lu, Lu, and Wang]{lu2021apriori}
Yulong Lu, Jianfeng Lu, and Min Wang.
\newblock A priori generalization analysis of the deep ritz method for solving high dimensional elliptic partial differential equations.
\newblock In \emph{Proceedings of Thirty Fourth Conference on Learning Theory}, volume 134 of \emph{Proceedings of Machine Learning Research}, pp.\  3196--3241. PMLR, 2021.

\bibitem[Luxburg \& Bousquet(2004)Luxburg and Bousquet]{luxburg2004distance}
Ulrike~von Luxburg and Olivier Bousquet.
\newblock Distance-based classification with {L}ipschitz functions.
\newblock \emph{Journal of Machine Learning Research}, 5\penalty0 (Jun):\penalty0 669--695, 2004.

\bibitem[Mac{\^e}do \& Castro(2010)Mac{\^e}do and Castro]{macêdo2010learning}
I.~Mac{\^e}do and R.~Castro.
\newblock \emph{Learning Divergence-free and Curl-free Vector Fields with Matrix-valued Kernels}.
\newblock IMPA, 2010.
\newblock URL \url{https://books.google.com/books?id=xgxCvwEACAAJ}.

\bibitem[Minakowski \& Richter(2023)Minakowski and Richter]{mina2023apriori}
P.~Minakowski and T.~Richter.
\newblock {A priori and a posteriori error estimates for the Deep Ritz method applied to the Laplace and Stokes problem}.
\newblock \emph{Journal of Computational and Applied Mathematics}, 421:\penalty0 114845, 2023.
\newblock ISSN 0377-0427.

\bibitem[Novak et~al.(2022)Novak, Sohl-Dickstein, and Schoenholz]{novak2022fast}
Roman Novak, Jascha Sohl-Dickstein, and Samuel~S Schoenholz.
\newblock Fast finite width neural tangent kernel.
\newblock In \emph{International Conference on Machine Learning}, pp.\  17018--17044. PMLR, 2022.

\bibitem[Pförtner et~al.(2024)Pförtner, Steinwart, Hennig, and Wenger]{pförtner2024physicsinformedgaussianprocessregression}
Marvin Pförtner, Ingo Steinwart, Philipp Hennig, and Jonathan Wenger.
\newblock Physics-informed {G}aussian process regression generalizes linear {PDE} solvers, 2024.
\newblock URL \url{https://arxiv.org/abs/2212.12474}.

\bibitem[Rahimi \& Recht(2007)Rahimi and Recht]{rahimi2007random}
Ali Rahimi and Benjamin Recht.
\newblock Random features for large-scale kernel machines.
\newblock In \emph{Advances in Neural Information Processing Systems}, volume~20, 2007.

\bibitem[Raissi et~al.(2019)Raissi, Perdikaris, and Karniadakis]{raissi2019physics}
Maziar Raissi, Paris Perdikaris, and George~E Karniadakis.
\newblock Physics-informed neural networks: A deep learning framework for solving forward and inverse problems involving nonlinear partial differential equations.
\newblock \emph{Journal of Computational physics}, 378:\penalty0 686--707, 2019.

\bibitem[Rasmussen \& Williams(2006)Rasmussen and Williams]{Rasmussen2006Gaussian}
Carl~Edward Rasmussen and Christopher K.~I. Williams.
\newblock \emph{Gaussian Processes for Machine Learning}.
\newblock The MIT Press, 2006.

\bibitem[Sahli~Costabal et~al.(2020)Sahli~Costabal, Yang, Perdikaris, Hurtado, and Kuhl]{sahli2020physics}
Francisco Sahli~Costabal, Yibo Yang, Paris Perdikaris, Daniel~E Hurtado, and Ellen Kuhl.
\newblock Physics-informed neural networks for cardiac activation mapping.
\newblock \emph{Frontiers in Physics}, 8:\penalty0 42, 2020.

\bibitem[Sakarvadia et~al.(2025)Sakarvadia, Hegazy, Totounferoush, Chard, Yang, Foster, and Mahoney]{FalsePromizeZeroShot_TR}
M.~Sakarvadia, K.~Hegazy, A.~Totounferoush, K.~Chard, Y.~Yang, I.~Foster, and M.~W. Mahoney.
\newblock The false promise of zero-shot super-resolution in machine-learned operators.
\newblock Technical Report Preprint: arXiv:2510.06646, 2025.

\bibitem[Schwarz(1978)]{schwarz1978estimating}
Gideon Schwarz.
\newblock Estimating the dimension of a model.
\newblock \emph{The annals of statistics}, pp.\  461--464, 1978.

\bibitem[Sirignano \& Spiliopoulos(2018)Sirignano and Spiliopoulos]{sirignano2018dgm}
Justin Sirignano and Konstantinos Spiliopoulos.
\newblock Dgm: A deep learning algorithm for solving partial differential equations.
\newblock \emph{Journal of computational physics}, 375:\penalty0 1339--1364, 2018.

\bibitem[Solin et~al.(2018)Solin, Kok, Wahlstrom, Schon, and Sarkka]{solin2018modeling}
Arno Solin, Manon Kok, Niklas Wahlstrom, Thomas~B. Schon, and Simo Sarkka.
\newblock Modeling and interpolation of the ambient magnetic field by {G}aussian processes.
\newblock \emph{Trans. Rob.}, 34\penalty0 (4):\penalty0 1112–1127, August 2018.
\newblock ISSN 1552-3098.
\newblock \doi{10.1109/TRO.2018.2830326}.

\bibitem[Steinwart \& Christmann(2008)Steinwart and Christmann]{steinwart2008support}
I.~Steinwart and A.~Christmann.
\newblock \emph{Support Vector Machines}.
\newblock Information Science and Statistics. Springer New York, 2008.
\newblock ISBN 9780387772424.
\newblock URL \url{https://books.google.com/books?id=HUnqnrpYt4IC}.

\bibitem[Trefethen(2019)]{trefethen2019approximation}
Lloyd~N. Trefethen.
\newblock \emph{Approximation Theory and Approximation Practice, Extended Edition}.
\newblock Society for Industrial and Applied Mathematics, Philadelphia, PA, 2019.

\bibitem[Wacker(2017)]{wacker2017laplace}
Philipp Wacker.
\newblock Laplace's method in {B}ayesian inverse problems.
\newblock \emph{arXiv preprint arXiv:1701.07989}, 2017.

\bibitem[Wendland(2004)]{Wendland_2004}
Holger Wendland.
\newblock \emph{Scattered Data Approximation}.
\newblock Cambridge Monographs on Applied and Computational Mathematics. Cambridge University Press, 2004.

\bibitem[Xu \& Darve(2020)Xu and Darve]{xu2020physics}
Kailai Xu and Eric Darve.
\newblock Physics constrained learning for data-driven inverse modeling from sparse observations.
\newblock \emph{arXiv preprint arXiv:2002.10521}, 2020.

\end{thebibliography}
\bibliographystyle{iclr2026_conference}

\appendix

\section{Reproducing Kernel Hilbert Spaces}
\label{sec:RKHS}

In this section, we provide a brief summary of reproducing kernel Hilbert spaces (RKHSs); for more details, we refer the interested reader to \citet{steinwart2008support}.
Let $H$ be a Hilbert space of functions $f:\mathcal{X} \to \mathbb{R}$ with an inner product $\langle \cdot ,\cdot \rangle_H$. $H$ is called a \emph{reproducing kernel Hilbert space} if for every $f \in H$, the evaluation functional $\iota_x: H \to \mathbb{R}$ defined for $x \in \mathcal{X}$ by $\iota_x f = f(x)$, is bounded, i.e., $|f(x)| \leq M_x \|f\|_H$ for some $M_x < +\infty$ and all $f \in H$. By the Riesz representation theorem, this implies that for any $x \in \mathcal{X}$, there exists $k_x \in H$ such that $f(x) = \langle f, k_x \rangle_H$. These elements $k_x$ are called the \emph{feature maps}. The reproducing kernel of the Hilbert space is given by
\[
k(x,y) = \langle k_x, k_y \rangle_H.
\]
The Moore-Aronszajn theorem \cite[Theorem 4.21]{steinwart2008support} implies that every positive-definite kernel induces a unique RKHS, so the kernel completely defines $H$. One can also deduce the class of functions contained in the RKHS from the kernel. For example, if the kernel is differentiable, this implies that elements of $H$ are also differentiable, as in the following proposition.

\begin{proposition}[Corollary 4.36 of \citet{steinwart2008support}]
\label{prop:DiffRKHS}
    Let $f \in H$. For $\alpha \leq \lceil s \rceil$ the derivative $\partial^\alpha f(x) \in C^0(\Omega)$ exists and admits the bound:
    $$
    |\partial^\alpha f(x)| \leq \|f\|_H \cdot (\partial^{\alpha}_1 \partial^\alpha_2 k(x, x))^{1/2}. 
    $$
    Hence $\Dcl f \in C^0(\Omega)$ for each $f \in H$. 
\end{proposition}

Furthermore, the image of $H$ under a linear operator is typically also an RKHS. The following proposition shows that the image of $H$ under a linear differential operator is an RKHS with a new kernel defined in terms of the operator.

\begin{proposition}\label{prop:differentiated-rkhs}
    The space $G \coloneqq \{g \in H : \exists \, f \in H \text{ with } g(x) = \Dcl f(x) \}$ endowed with the norm 
    \begin{align*}
        \|g\|_G = \inf_{f : \Dcl f = g} \|f \|_H
    \end{align*}
    is the unique RKHS associated with the kernel 
    $$g(x, y) \coloneqq (\Dcl \otimes \Dcl) k(x, x').$$ Moreover, $g(x, x') = \langle g_x, g_{x'} \rangle_H$ where the feature map $g_x \in H$ is given by $g_x(x') = (\Dcl \otimes \mathsf{Id}) k(x, x')$. 
\end{proposition} 

\begin{proof}
    It is sufficient to consider the case $\Dcl f(x) = c_\alpha(x) \partial^{\alpha}f(x)$ for some $\alpha \in \Z$, as the general result follows from summing over terms of $\Dcl f$. 
    For any $\alpha$, it holds $y \mapsto \partial^\alpha_x k(x, y) \in H$ by repeated application of Lemma 4.34 of \citet{steinwart2008support}. 
    Hence, 
    $$
    \|(\Dcl \otimes \mathsf{Id})(x, \cdot) \|_H = |c_\alpha(x)| \cdot \|(\partial^\alpha \otimes \mathsf{Id}) k\|_H < \infty,
    $$
    and 
    \begin{align*}
    \langle f, \Dcl^* k_x \rangle_H & = c_\alpha(x) \langle f, (\partial^\alpha \otimes \mathsf{Id})k(x, \cdot) \rangle_H 
 \\
 & = c_\alpha(x) \partial^\alpha_x \langle f, k_x \rangle_{H} \\
 & = \Dcl f(x) ,
    \end{align*}
    by the chain rule.
\end{proof}

\begin{proposition}
\label{prop:NewRKHS}
The graph $\{(f,\mathcal{A}f)\,:\, f\in H\}$ is a RKHS $H_{\Acl}$ with norm
\[
\|(f,\mathcal{A}f)\|_{H_{\Acl}} = \|f\|_H
\]
and a multi-valued reproducing kernel $k_{\Acl}$ given by
\[
k_{\Acl}((x,z),(x',z')) = \begin{bmatrix}
k(x,x') & (\mathsf{Id} \otimes \Acl) k(x',z) \\ (\mathsf{Id} \otimes \Acl)k(x,z') & (\Acl \otimes \Acl) k(z,z')
\end{bmatrix} .
\]
\end{proposition} 

Proposition \ref{prop:differentiated-rkhs} is sufficient to prove Proposition \ref{prop:NewRKHS}. To see this, note that the space $H_\Acl=\{(f,\Acl f) \in H \oplus H \, : \, f \in H\}$ can be equipped with the inner product acting only on the first coordinate: $\langle (f,\Acl f), (g, \Acl g)\rangle_{H_\Acl}$ = $\langle f, g\rangle_H$. The evaluation functionals are guaranteed to be bounded by Proposition \ref{prop:differentiated-rkhs}, and the form of the kernel follows. 

Here, we let $W = \mbox{diag}(w_i)_{i=1}^m$, so that $W^{1/2}$ can be interpreted as the elementwise square~root.
\begin{theorem}[Representer Theorem]
\label{prop:representer-theorem}
The minimizer $\hat{f} \coloneqq \arg \inf_{f \in H} L_{m,n}(f)$ satisfies
    \begin{align}\label{eqn:pinn-rkhs-span}
        \hat{f} \in \mathrm{span} \left[ \{k(\cdot, x_i) \}_{i=1}^n \cup \{(\mathsf{Id} \otimes \Acl)k(\cdot, z_j)\}_{j=1}^m \right],
    \end{align}
and for $\hat{f} = \sum_{i=1}^n \alpha_i k(\cdot,x_i) + \sum_{j=1}^m \beta_j (\mathsf{Id}\otimes\mathcal{A})k(\cdot,z_j)$, the coefficients $(\alpha, \beta) \in \R^{m+n}$ minimize \begin{equation}
\frac{1}{\gamma n} \left\| 
Y - K_{xx}\, \alpha - H_{xz}\, \beta
\right\|^2_2  + 
\frac{1}{\rho} \|W^{1/2}(H_{xz}^\top\,  \alpha - G_{zz} \beta )\|_2^2 + 
\frac{1}{2\eta}\label{eqn:representer-theorem}
\begin{bmatrix}
    \alpha \\ \beta
\end{bmatrix}^\top
\begin{bmatrix}
    K_{xx} & H_{xz} \\
    H_{xz}^\top & G_{zz}
\end{bmatrix}
\begin{bmatrix}
    \alpha \\ \beta
\end{bmatrix} ,
\end{equation}
which is now equal to $L_{m,n}(\hat{f})$, where $[K_{xx}]_{i,j} = k(x_i, x_j)$, $[H_{xz}]_{ij} = (\mathsf{Id} \otimes \Dcl)k(x_i, z_j)$, and $[G_{zz}]_{ij} = (\Dcl \otimes \Dcl) k(z_i, z_j)$. 
\end{theorem}

\begin{proof}[Proof of Theorem \ref{prop:representer-theorem}]
    The property in \eqref{eqn:pinn-rkhs-span} follows from the fact that the $\|\cdot\|_H$ projection of any $f \in H$ onto the span does not affect the values of $f(x_i)$ nor $\Dcl f(z_j))$. 
    By the representer property for $H$,
    \begin{align*}
        \langle k_{x_i}, g_{z_j} \rangle_H = \langle k_{x_i}(\cdot) , (\mathsf{Id} \otimes \Dcl) k(\cdot, z_j)\rangle_H = (\mathsf{Id} \otimes \Dcl)k(x_i, z_j) ,
    \end{align*}
    and by Proposition \ref{prop:differentiated-rkhs}, $\langle g_{z_i}, g_{z_j} \rangle = (\Dcl \otimes \Dcl)k(z_i, z_j)$. 
    Plugging these identities into $\|\hat{f}\|_H^2$ yields 
    \begin{align*}
    \|\hat{f}\|_H^2 = 
        \begin{bmatrix} \alpha \\ \beta \end{bmatrix}^\top \begin{bmatrix}
            K_{xx} & H_{xz} \\ H_{xz}^\top & G_{zz}
        \end{bmatrix}
        \begin{bmatrix} \alpha \\ \beta \end{bmatrix}.
    \end{align*} 
    Observe that \eqref{eqn:representer-theorem} is just \eqref{eqn:pinn-rkhs-span} rewritten in terms of $\alpha, \beta$. 
\end{proof}

\section{Fredholm Determinants}
\label{sec:FredholmApp}

In this section, we describe in greater detail the connection between PILE and Fredholm Determinants.

Let $H$ be a separable Hilbert space with an inner product $\langle\cdot,\cdot\rangle_H$, and let $\mathcal{K}(H)$ denote the space of compact linear operators $A:H \to H$. The spectrum
\[
\mathrm{spec}(A) = \{\lambda \, : \, \mbox{ker}(\lambda I - C) \neq \{0\}\},
\]
of any $A \in \mathcal{K}(H)$ is countable and can accumulate only at zero. Hence, there is a (possibly infinite) sequence of eigenvalues $\{\lambda_n(A)\}_n$. For any $A \in \mathcal{K}(H)$, let
\[
\sigma_1(A) \geq \sigma_2(A) \geq \cdots > 0,
\]
denote the ordered singular values of $A$, defined as the square root of the eigenvalues of $A^* A$.
\begin{definition}
A compact operator $A \in \mathcal{K}(H)$ is \emph{trace class} if $\sum_n \sigma_n(A) < +\infty$. 
\end{definition}
By Holder's inequality, for any trace class operator $A$, the Schatten norm satisfying
\[
\|A\|_p^p = \sum_n \sigma_n(A)^p,\qquad p \geq 1,
\]
is finite. Using the definition to verify whether an operator is trace class is a near-impossible task in general.
Fortunately, the following theorem provides conditions that guarantee in our kernel setting that all operators are trace class. 
\begin{theorem}
Let $k$ be a symmetric positive-definite kernel on $\Omega \times \Omega$ such that $x \mapsto k(x, x) \in L^2(\Omega)$ for $\Omega \subseteq \mathbb{R}^d$. The integral operator $K:L^2(\Omega) \to L^2(\Omega)$ defined by
\[
(K f)(x) = \int_{\Omega} k(x, y) f(y) \dd y,\qquad \text{ is trace class.}
\]
\end{theorem}
We are now ready to define the Fredholm determinant of an operator.
\begin{definition}
The \emph{Fredholm determinant} of a trace class operator $A$ is given for $\rho > 0$ by
\[
\det(I + \rho A) = \prod_n (1 + \rho s_n(A)).
\]
\end{definition}
It is straightforward to verify that the Fredholm determinant can only be defined for a trace class operator, since \cite[p. 232]{knopp2013theorie}
\[
1 + \rho \|A\|_1 \leq \det(I + \rho A) \leq \exp(\rho \|A\|_1).
\]
Note that for $H = \mathbb{R}^m$, the Fredholm determinant reduces to the standard determinant, as a bounded linear operator $A:\mathbb{R}^m \to \mathbb{R}^m$ can be represented as a matrix $A_m$ over the basis elements $\{e_i\}_{i=1}^m$,~and
\[
\det(I + \rho A) = \prod_{n=1}^m (1 + \rho s_n(A)) = \det(I + \rho A_m).
\]
In the case of integral operators, the Fredholm determinant is expressed in terms of the infinite series \cite[eqn. (3.7)]{bornemann2010numerical}
\[
\det(I + \rho K) = 1 + \sum_{n=1}^{\infty} \frac{\rho^n}{n!} \int_{\Omega^n} \det(k(x_i,x_j))_{i,j=1}^n \dd x_1 \cdots \dd x_n.
\]

\begin{proof}[Proof of Theorem \ref{thm:Fredholm}]
Starting from the PILE score,
\begin{align*}
m\mathfrak{P}_{m,n}&=\log\det(G_{zz}+\eta\rho W^{-1})+m\log(2\pi\eta)\\
&=\log\det(W^{1/2}G_{zz}W^{1/2}+\eta\rho I)+m\log(2\pi\eta)-\sum_{i=1}^{m}\log w_{i}\\
&=\log\det(I+W^{1/2}(\eta\rho)^{-1}G_{zz}W^{1/2})+C_{m},
\end{align*}
and consequently,
\[
m\mathfrak{P}_{m,n} - C_m = \log \det(I + W^{1/2} ((\mathcal{A}\otimes\mathcal{A})k(z_i,z_j))_{i,j=1}^m W^{1/2}).
\]
The result now follows from \cite[Theorem 6.1]{bornemann2010numerical}.
\end{proof}




\end{document}